\documentclass[runningheads]{llncs}
\usepackage{graphicx}
\usepackage{amsxtra}
\usepackage[caption = false]{subfig}
\usepackage{tabularx}
\usepackage{makecell}
\usepackage{booktabs}
\usepackage{multirow}
\usepackage{cite}
\usepackage{tikz}
\usepackage{comment}
\usepackage{amsmath,amssymb} % define this before the line numbering.
\usepackage{color}

\usepackage[pagebackref=true,breaklinks=true,letterpaper=true,colorlinks,bookmarks=false]{hyperref}

\newcommand{\etal}{\textit{et al}.}
\newcommand{\resitem}[1]{\item #1}

\begin{document}
\newcolumntype{L}[1]{>{\raggedright\arraybackslash}p{#1}}
\newcolumntype{C}[1]{>{\centering\arraybackslash}p{#1}}
\newcolumntype{R}[1]{>{\raggedleft\arraybackslash}p{#1}}

\pagestyle{headings}
\mainmatter
\def\ECCVSubNumber{2095}  % Insert your submission number here

\title{BMBC:Bilateral Motion Estimation with Bilateral Cost Volume for Video Interpolation} % Replace with your title

\begin{comment}
\titlerunning{ECCV-20 submission ID \ECCVSubNumber}
\authorrunning{ECCV-20 submission ID \ECCVSubNumber}
\author{Anonymous ECCV submission}
\institute{Paper ID \ECCVSubNumber}
\end{comment}
%******************

%\begin{comment}
\titlerunning{BMBC:Bilateral Motion Estimation \ldots~for Video Interpolation}
\author{Junheum Park\inst{1}\orcidID{0000-0002-9094-128X} \and
Keunsoo Ko\inst{1}\orcidID{0000-0003-0203-4530} \and \\
Chul Lee\inst{2}\orcidID{0000-0001-9329-7365} \and
Chang-Su Kim\inst{1}\orcidID{0000-0002-4276-1831}}
\authorrunning{J. Park \etal}
\institute{School of Electrical Engineering, Korea University, Seoul, Korea\and
Department of Multimedia Engineering, Dongguk University, Seoul, Korea\\
\email{\{jhpark,ksko\}@mcl.korea.ac.kr, chullee@dongguk.edu, changsukim@korea.ac.kr}}
%\end{comment}
%******************
\maketitle

\begin{abstract}
Video interpolation increases the temporal resolution of a video sequence by synthesizing intermediate frames between two consecutive frames. We propose a novel deep-learning-based video interpolation algorithm based on bilateral motion estimation. First, we develop the bilateral motion network with the bilateral cost volume to estimate bilateral motions accurately. Then, we approximate bi-directional motions to predict a different kind of bilateral motions. We then warp the two input frames using the estimated bilateral motions. Next, we develop the dynamic filter generation network to yield dynamic blending filters. Finally, we combine the warped frames using the dynamic blending filters to generate intermediate frames. Experimental results show that the proposed algorithm outperforms the state-of-the-art video interpolation algorithms on several benchmark datasets. The source codes and pre-trained models are available at \url{https://github.com/JunHeum/BMBC}.
\end{abstract}

\section{Introduction}

A low temporal resolution causes aliasing, yields abrupt motion artifacts, and degrades the video quality. In other words, the temporal resolution is an important factor affecting video quality. To enhance temporal resolutions, many video interpolation algorithms~\cite{choi2007motion,jeong2013texture,bao2018memc,bao2019dain,jiang2018slomo,liu2017dvf,liu2019cyclicgen,long2016learning,niklaus2017adaconv,niklaus2017sepconv,niklaus2018ctx} have been proposed, which synthesize intermediate frames between two actual frames. These algorithms are widely used in applications, including visual quality enhancement~\cite{xue2019toflow}, video compression~\cite{Lu2018compression}, slow-motion video generation~\cite{jiang2018slomo}, and view synthesis~\cite{Flynn2016viewsynthesis}. However, video interpolation is challenging due to diverse factors, such as large and nonlinear motions, occlusions, and variations in lighting conditions. Especially, to generate a high-quality intermediate frame, it is important to estimate motions or optical flow vectors accurately.

Recently, with the advance of deep-learning-based optical flow methods~\cite{dosovitskiy2015flownet,ilg2017flownet2,ranjan2017spynet,sun2018pwc}, flow-based video interpolation algorithms~\cite{bao2018memc,bao2019dain,jiang2018slomo} have been developed, yielding reliable interpolation results. Niklaus \etal~\cite{niklaus2018ctx} generated intermediate frames based on the forward warping. However, the forward warping may cause interpolation artifacts because of the hole and overlapped pixel problems. To overcome this, other approaches leverage the backward warping. To use the backward warping, intermediate motions should be obtained.
Various video interpolation algorithms~\cite{choi2007motion,jiang2018slomo,bao2018memc,bao2019dain,niklaus2018ctx,xue2019toflow,liu2019cyclicgen} based on the bilateral motion estimation approximate these intermediate motions from optical flows between two input frames. However, this approximation may degrade video interpolation results.

In this work, we propose a novel video interpolation network, which consists of the bilateral motion network and the dynamic filter generation network. First, we predict six bilateral motions: two from the bilateral motion network and the other four through optical flow approximation. In the bilateral motion network, we develop the bilateral cost volume to facilitate the matching process. Second, we extract context maps to exploit rich contextual information. We then warp the two input frames and the corresponding context maps using the six bilateral motions, resulting in six pairs of warped frame and context map. Next, these pairs are used to generate dynamic blending filters. Finally, the six warped frames are superposed by the blending filters to generate an intermediate frame. Experimental results demonstrate that the proposed algorithm outperforms the state-of-the-art video interpolation algorithms~\cite{bao2018memc,bao2019dain,long2016learning,niklaus2017sepconv,xue2019toflow} meaningfully on various benchmark datasets.

This work has the following major contributions:
\begin{itemize}
    \setlength\itemsep{0.5em}
    \resitem{We develop a novel deep-learning-based video interpolation algorithm based on the bilateral motion estimation.}
    \resitem{We propose the bilateral motion network with the bilateral cost volume to estimate intermediate motions accurately.}
    \resitem{The proposed algorithm performs better than the state-of-the-art algorithms on various benchmark datasets.}
\end{itemize}

\section{Related Work}

\subsection{Deep-learning-based video interpolation}
The objective of video interpolation is to enhance a low temporal resolution by synthesizing intermediate frames between two actual frames. With the great success of CNNs in various image processing and computer vision tasks, many deep-learning-based video interpolation techniques have been developed. Long \etal~\cite{long2016learning} developed a CNN, which takes a pair of frames as input and then directly generates an intermediate frame. However, their algorithm yields severe blurring since it does not use a motion model. In~\cite{Meyer2018phase}, PhaseNet was proposed using the phase-based motion representation. Although it yields robust results to lightning changes or motion blur, it may fail to faithfully reconstruct detailed texture. In~\cite{niklaus2017adaconv,niklaus2017sepconv}, Niklaus \etal~proposed kernel-based methods that estimate an adaptive convolutional kernel for each pixel. The kernel-based methods produce reasonable results, but they cannot handle motions larger than a kernel's size.

To exploit motion information explicitly, flow-based algorithms have been developed. Niklaus and Liu~\cite{niklaus2018ctx} generated an intermediate frame from two consecutive frames using the forward warping. However, the forward warping suffers from holes and overlapped pixels. Therefore, most flow-based algorithms are based on backward warping. In order to use backward warping, intermediate motions (\textit{i.e.}~motion vectors of intermediate frames) should be estimated. Jiang \etal~\cite{jiang2018slomo} estimated optical flows and performed bilateral motion approximation to predict intermediate motions from the optical flows. Bao \etal~\cite{bao2018memc} approximated intermediate motions based on the flow projection. However, large errors may occur when two flows are projected onto the same pixel. In~\cite{bao2019dain}, Bao \etal~proposed an advanced projection method using the depth information. However, the resultant intermediate motions are sensitive to the depth estimation performance.
To summarize, although the backward warping yields reasonable video interpolation results, its performance degrades severely when intermediate motions are unreliable or erroneous. To solve this problem, we propose the bilateral motion network to estimate intermediate motions directly.

\subsection{Cost volume}
A cost volume records similarity scores between two data. For example, in pixel-wise matching between two images, the similarity is computed between each pixel pair: one in a reference image and the other in a target image. Then, for each reference pixel, the target pixel with the highest similarity score becomes the matched pixel. The cost volume facilitates this matching process. Thus, the optical flow estimation techniques in~\cite{dosovitskiy2015flownet, jiaxu2017dcflow,ilg2017flownet2,sun2018pwc} are implemented using cost volumes. In~\cite{dosovitskiy2015flownet,ilg2017flownet2,jiaxu2017dcflow}, a cost volume is computed using various features of two video frames, and optical flow is estimated using the similarity information in the cost volume through a CNN. Sun \etal~\cite{sun2018pwc} proposed a partial cost volume to significantly reduce the memory requirement while improving the motion estimation accuracy based on a reduced search region. In this work, we develop a novel cost volume, called bilateral cost volume, which is different from the conventional volumes in that its reference is an intermediate frame to be interpolated, instead of one of the two input frames.

\section{Proposed Algorithm}

\begin{figure}[t]
    \setlength{\belowcaptionskip}{-10pt}
    \centering
    \includegraphics[width=12cm]{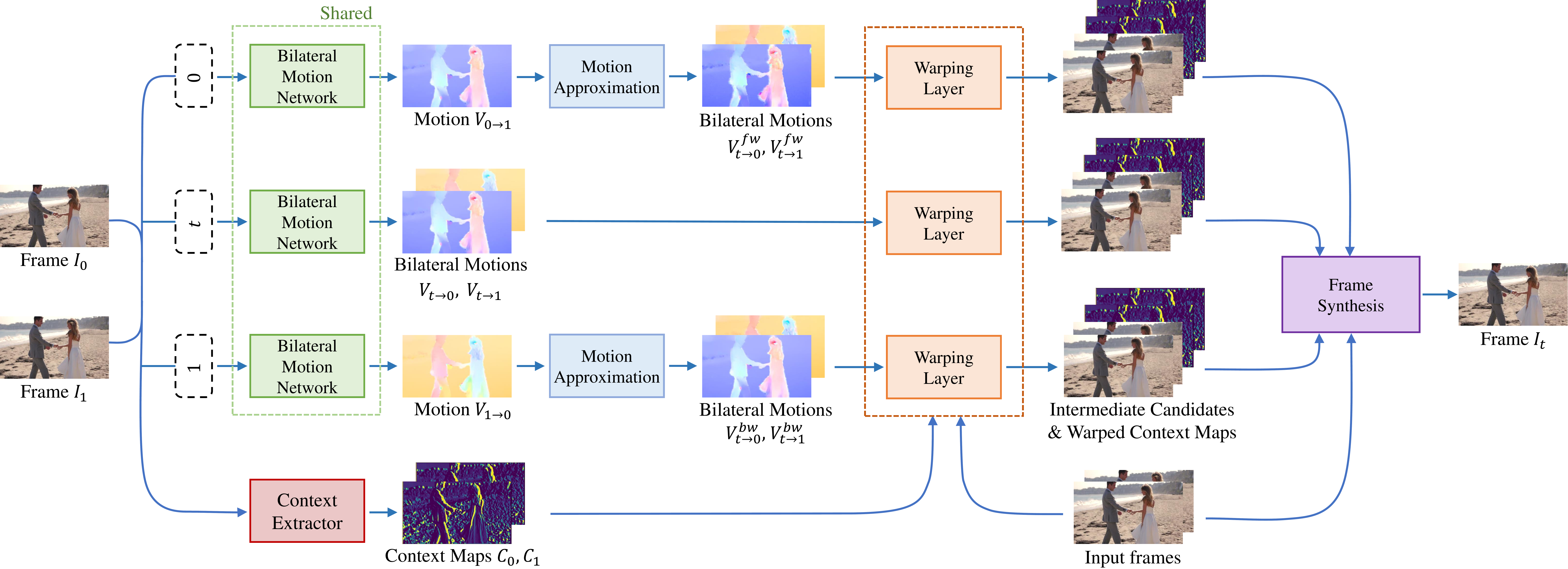}
    \caption{An overview of the proposed video interpolation algorithm.}
    % We first estimate bi-directional motions $V_{0 \rightarrow 1}$ and $V_{1 \rightarrow 0}$ and bilateral motions $V_{t \rightarrow 0}$ and $V_{0 \rightarrow t}$. Four additional bilateral motions are approximated using the bi-directional motions. Context maps are extracted from the input frames. Then, each bilateral motion is used to generate a warped frame, which is a candidate of the intermediate frame. A stack of the six intermediate candidates and the two input frames are fed into the dynamic filter network to generate local blending filters. Finally, the intermediate frame is synthesized by applying the blending filters to the intermediate candidates.
    \label{fig:overview}
\end{figure}

Fig. \ref{fig:overview} is an overview of the proposed algorithm that takes successive frames $I_{0}$ and $I_{1}$ as input and synthesizes an intermediate frame $I_{t}$ at $t \in (0,1)$ as output. First, we estimate two `bilateral' motions $V_{t \rightarrow 0}$ and $V_{0 \rightarrow t}$ between the input frames. Second, we estimate `bi-directional' motions $V_{0 \rightarrow 1}$ and $V_{1 \rightarrow 0}$ between $I_{0}$ and $I_{1}$ and then use these motions to approximate four further bilateral motions. Third, the pixel-wise context maps $C_0$ and $C_1$ are extracted from $I_0$ and $I_1$. Then, the input frames and corresponding context maps are warped using the six bilateral motions. Note that, since the warped frames become multiple candidates of the intermediate frame, we refer to each warped frame as an intermediate candidate. The dynamic filter network then takes the input frames, and the intermediate candidates with the corresponding warped context maps to generate the dynamic filters for aggregating the intermediate candidates. Finally, the intermediate frame $I_{t}$ is synthesized by applying the blending filters to the intermediate candidates.

%%%%%%%%%%%
\subsection{Bilateral motion estimation}

Given the two input frames $I_0$ and $I_1$, the goal is to predict the intermediate frame $I_{t}$ using motion information. However, it is impossible to directly estimate the intermediate motion between the intermediate frame $I_{t}$ and one of the input frames $I_0$ or $I_1$ because there is no image information of $I_{t}$. To address this issue, we assume linear motion between successive frames. Specifically, we attempt to estimate the backward and forward motion vectors $V_{t \rightarrow 0}({\bf x})$ and $V_{t \rightarrow1}(\bf x)$ at $\bf x$, respectively, where $\bf x$ is a pixel location in $I_{t}$. Based on the linear assumption, we have $V_{t \rightarrow 0}({\bf x}) = -\frac{t}{1-t} \times V_{t \rightarrow 1}({\bf x})$.

\begin{figure}[t]
    \centering
    \includegraphics[width=11cm]{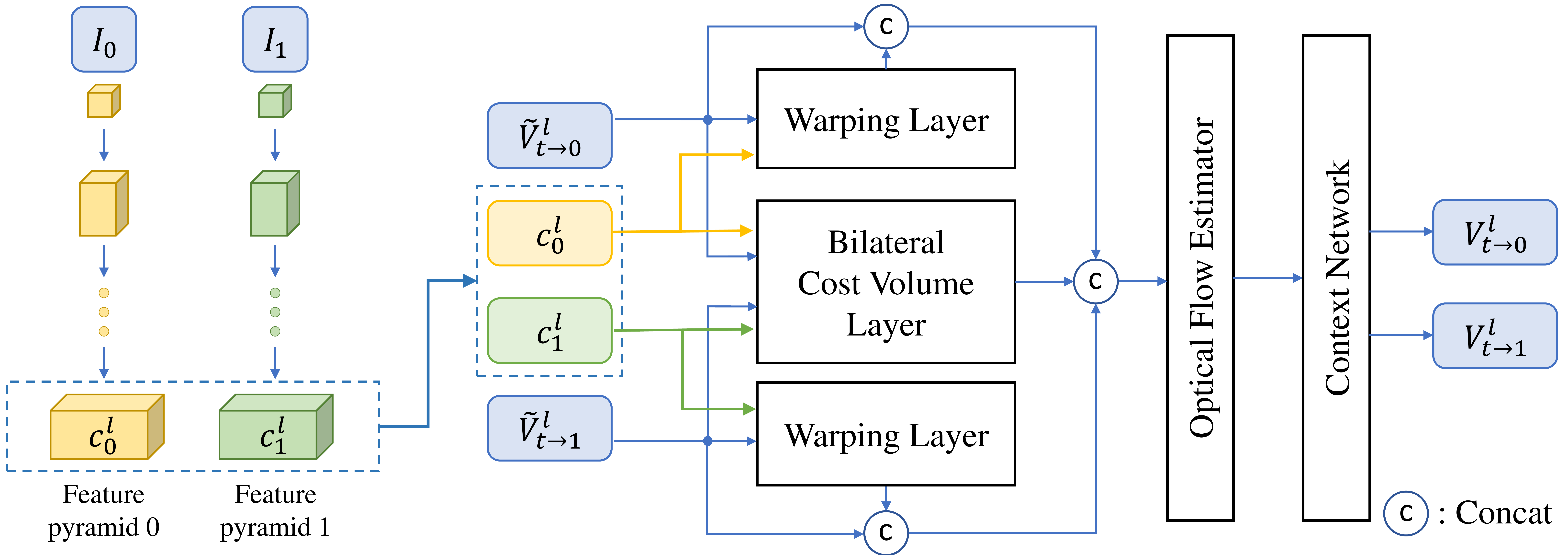}
    \caption
    {
        The architecture of the bilateral motion network: The feature maps $c^l_0$ and $c^l_1$ of the previous and following frames $I_0$ and $I_1$ at the $l$th level and the up-sampled motion fields $\widetilde{V}^l_{t \rightarrow 0}$ and $\widetilde{V}^l_{t \rightarrow 1}$ estimated at the ($l- 1$)th level are fed into the CNN to generate the motion fields $V^l_{t \rightarrow 0}$ and $V^l_{t \rightarrow 1}$ at the $l$th level.
    }
  \label{fig:Bilateral Motion Network}
\end{figure}

We develop a CNN to estimate bilateral motions $V_{t \rightarrow 0}$ and $V_{t \rightarrow 1}$ using $I_0$ and $I_1$. To this end, we adopt an optical flow network, PWC-Net~\cite{sun2018pwc}, and extend it for the bilateral motion estimation. Fig.~\ref{fig:Bilateral Motion Network} shows the key components of the modified PWC-Net. Let us describe each component subsequently.

\subsubsection{Warping layer:} The original PWC-Net uses the previous frame $I_0$ as a reference and the following frame $I_1$ as a target. On the other hand, the bilateral motion estimation uses the intermediate frame $I_t$ as a reference, and the input frames $I_0$ and $I_1$ as the target. Thus, whereas the original PWC-Net warps the feature $c^l_1$ of $I_1$ toward the feature $c^l_0$ of $I_0$, we warp both features $c^l_0$ and $c^l_1$ toward the intermediate frame, leading to $c^l_{0 \rightarrow t}$ and $c^l_{1 \rightarrow t}$, respectively. We employ the spatial transformer networks~\cite{Jaderberg2015STN} to achieve the warping. Specifically, a target feature map $c_{\rm tgt}$ is warped into a reference feature map $c_{\rm ref}$ using a motion vector field by
\begin{equation}\label{eq:warping}
  c^w_{\rm ref}({\bf x}) = c_{\rm tgt}\big({\bf x} + V_{\rm ref \rightarrow tgt}({\bf x})\big)
\end{equation}
where $V_{\rm ref\rightarrow tgt}$ is the motion vector field from the reference to the target.
%In the bilateral motion network, we warp features of each input frame into intermediate frame using bilateral motion.
%All of warping layer in proposed algorithm denotes (\ref{eq:warping}).

\subsubsection{Bilateral cost volume layer:}

\begin{figure}[t]
    \setlength{\belowcaptionskip}{-10pt}
    \centering
    \includegraphics[width=10.5cm]{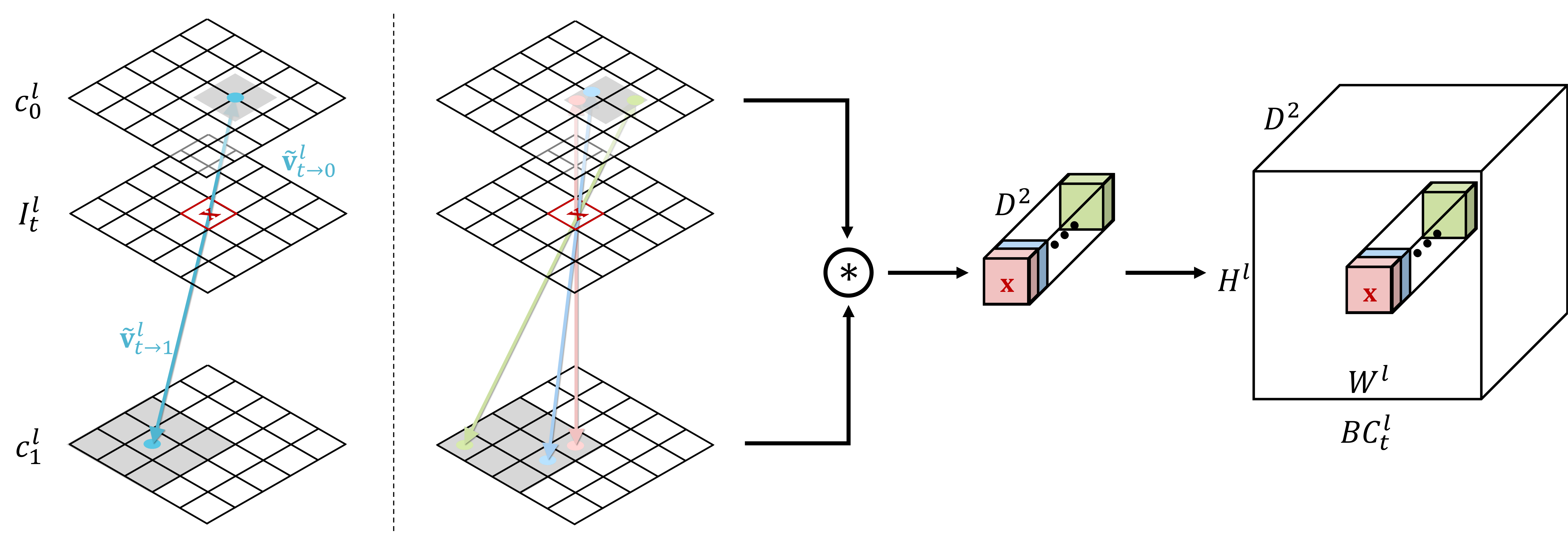}
    \caption{Illustration of the bilateral cost volume layer for a specific time $t$.}
    \label{fig:Bilateral cost volume layer}
\end{figure}

A cost volume has been used to store the matching costs associating with a pixel in a reference frame with its corresponding pixels in a single target frame~\cite{hosni2013fastcostvolume, dosovitskiy2015flownet, jiaxu2017dcflow, sun2018pwc}. However, in the bilateral motion estimation, because the reference frame does not exist and should be predicted from two target frames, the conventional cost volume cannot be used. Thus, we develop a new cost volume for the bilateral motion estimation, which we refer to as the bilateral cost volume.

Fig.~\ref{fig:Bilateral cost volume layer} illustrates the proposed bilateral cost volume generation that takes the features $c^l_0$ and $c^l_1$ of the two input frames and the up-sampled bilateral motion fields $\widetilde{V}^l_{t \rightarrow 0}$ and $\widetilde{V}^l_{t \rightarrow 1}$ estimated at the ($l-1$)th level. Let $\bf x$ denote a pixel location in the intermediate frame $I^l_t$. Then, we define the matching cost as the bilateral correlation between features $c^l_0$ and $c^l_1$, indexed by the bilateral motion vector that passes through $\bf x$, given by
\begin{equation}
    {BC^l_{t}(\mathbf{x},\mathbf{d})} = {c^l_0}(\mathbf{x}+\widetilde{V}^l_{\rm {t}\rightarrow{0}}(\mathbf{x})-2t\times\mathbf{d})^T {c^l_1}(\mathbf{x}+\widetilde{V}^l_{\rm {t}\rightarrow{1}}(\mathbf{x})+2(1-t)\times\mathbf{d})
    \label{eq:bilateral_cost_volume}
\end{equation}
where $\mathbf{d}$ denotes the displacement vector within the search window $\mathcal{D}=[-d,d]\times[-d,d]$.
%$\widetilde{\mathbf{v}}^l_{\rm {t}\rightarrow{0}}$,$\widetilde{\mathbf{v}}^l_{\rm {t}\rightarrow{1}}$ are $\times 2$ upsampled bilateral motions from $l-1$th level.
Note that we compute only $|\mathcal{D}| = D^2$ bilateral correlations to construct a partial cost volume, where $D=2d+1$. In the $L$-level pyramid architecture, a one-pixel motion at the coarsest level corresponds to $2^{L-1}$ pixels at the finest resolution. Thus, the search range $D$ of the bilateral cost volume can be set to a small value to reduce the memory usage. The dimension of the bilateral cost volume at the $l$th level is $D^2\times H^l\times W^l$, where $H^l$ and $W^l$ denote the height and width of the $l$th level features, respectively. Also, the up-sampled bilateral motions $\widetilde{V}^l_{t \rightarrow 0}$ and $\widetilde{V}^l_{t \rightarrow 1}$ are set to zero at the coarsest level.

Most conventional video interpolation algorithms generate a single intermediate frame at the middle of two input frames, {\it i.e.}~$t=0.5$. Thus, they cannot yield output videos with arbitrary frame rates.
A few recent algorithms~\cite{jiang2018slomo, bao2019dain} attempt to interpolate intermediate frames at arbitrary time instances $t\in(0,1)$. However, because their approaches are based on the approximation, as the time instance gets far from either of the input frames, the quality of the interpolated frame gets worse. On the other hand, the proposed algorithm takes into account the time instance $t \in [0,1]$ during the computation of the bilateral cost volume in (\ref{eq:bilateral_cost_volume}). Also, after we train the bilateral motion network with the bilateral cost volume, we can use the shared weights to estimate the bilateral motions at an arbitrary $t \in [0,1]$. In the extreme cases $t=0$ or $t=1$, the bilateral cost volume becomes identical to the conventional cost volume in~\cite{dosovitskiy2015flownet, chen2016fullflow, jiaxu2017dcflow, ilg2017flownet2, sun2018pwc}, which is used to estimate the bi-directional motions $V_{0 \rightarrow 1}$ and $V_{1 \rightarrow 0}$ between input frames.

%In addition, as shown in Fig.~\ref{fig:Bilateral Motion Network}, the proposed bilateral cost volume layer takes the features $c^l_0$ and $c^l_1$, instead of the warped version in the original PWC-Net, thereby suppressing geometric distortions more effectively.

\begin{figure*}[t]
    \setlength{\belowcaptionskip}{-10pt}
	\centering

	\subfloat []{\includegraphics[height=1.5cm]{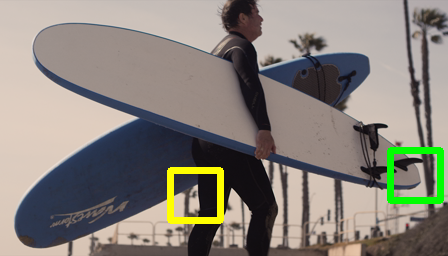}}\!
	\subfloat []{\includegraphics[height=1.5cm]{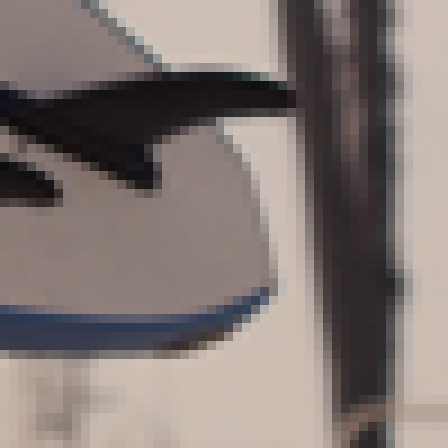}}\!
	\subfloat []{\includegraphics[height=1.5cm]{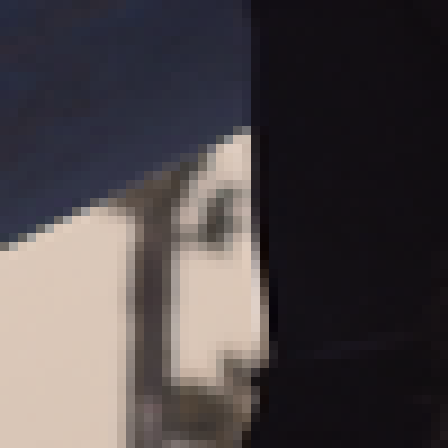}}\!
	\subfloat []{\includegraphics[height=1.5cm]{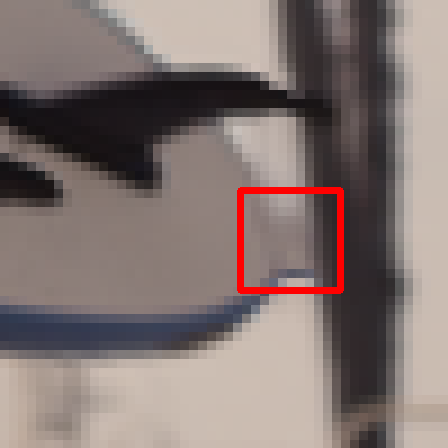}}\!
	\subfloat []{\includegraphics[height=1.5cm]{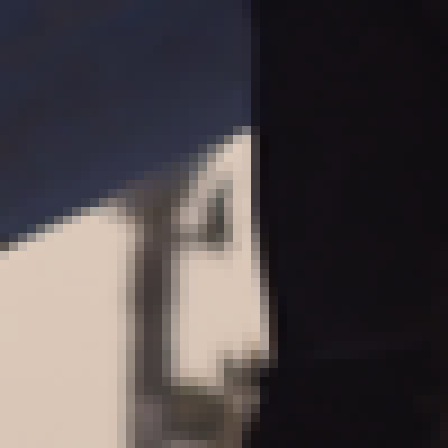}}\!
	\subfloat []{\includegraphics[height=1.5cm]{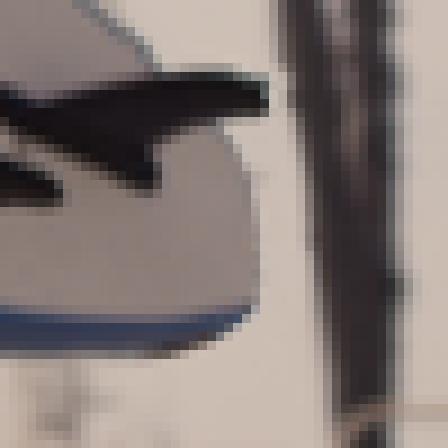}}\!
	\subfloat []{\includegraphics[height=1.5cm]{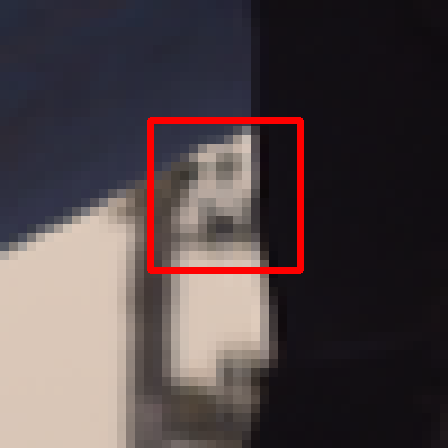}}\\[-0.5em]

    \caption
    {
        Comparison of interpolation results obtained by the bilinear motion estimation and the motion approximation: (a) ground-truth intermediate frame; (b), (c) enlarged parts for the green and yellow squares in (a); (d), (e) interpolation results using the bilateral motion estimation; (f), (g) those using the motion approximation. The red squares in (d) and (g) contain visual artifacts caused by motion inaccuracies.
    }
    \label{fig:Intermediate candidates}
\end{figure*}

\subsection{Motion approximation}

Although the proposed bilateral motion network effectively estimates motion fields $V_{t \rightarrow 0}$ and $V_{t \rightarrow 1}$ from the intermediate frame at $t$ to the previous and following frames, it may fail to find accurate motions, especially at occluded regions. For example, Fig.~\ref{fig:Intermediate candidates}(d) and (e) show that the interpolated regions, reconstructed by the bilateral motion estimation, contain visual artifacts. To address this issue and improve the quality of an interpolated frame, in addition to the bilateral motion estimation, we develop an approximation scheme to predict a different kind of bilateral motions $V_{t \rightarrow 0}$ and $V_{t \rightarrow 1}$ using the bi-directional motions $V_{0 \rightarrow 1}$ and $V_{1 \rightarrow 0}$ between the two input frames.

\begin{figure*}[t]
    \setlength{\belowcaptionskip}{-10pt}
    \centering
    \subfloat[$V_{0 \rightarrow 1}$ and $V_{1 \rightarrow 0}$]{\includegraphics[width=3.8cm]{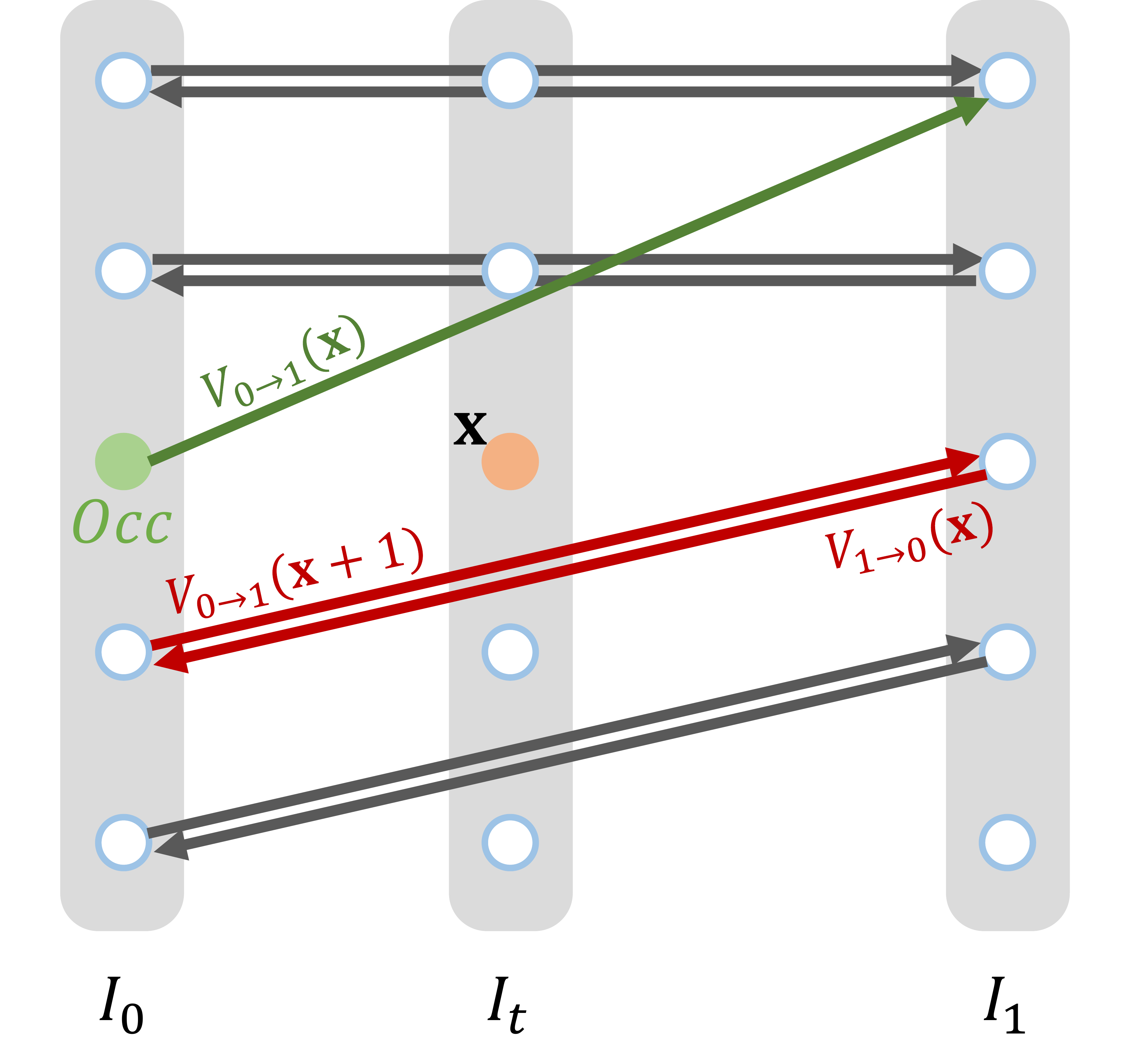}}\;
    \subfloat[$V^{fw}_{t \rightarrow 0}$ and $V^{fw}_{t \rightarrow 1}$]{\includegraphics[width=3.8cm]{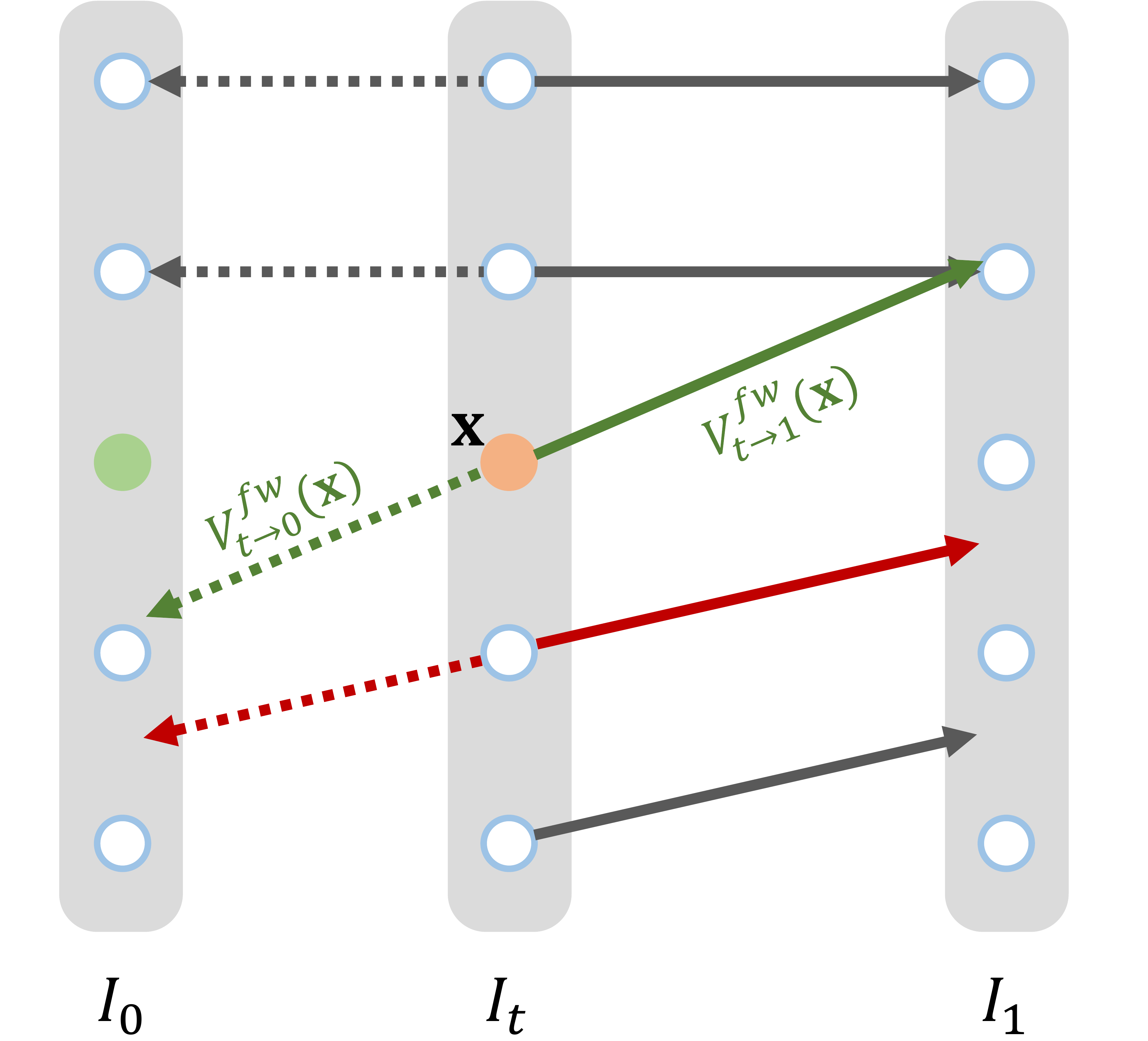}}\;
    \subfloat[$V^{bw}_{t \rightarrow 0}$ and $V^{bw}_{t \rightarrow 1}$]{\includegraphics[width=3.8cm]{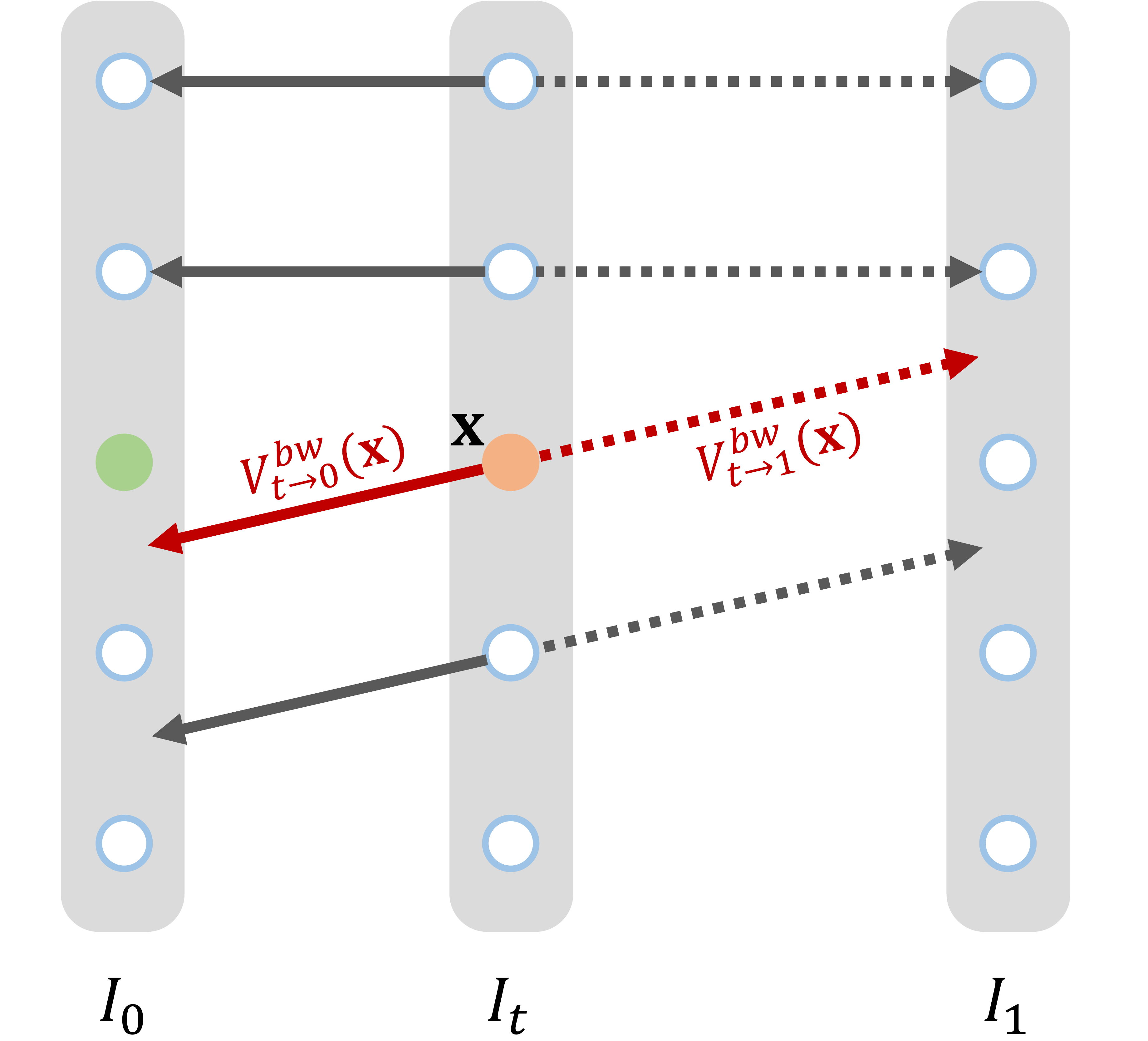}}\\[-0.3em]

    \caption
    {Motion approximation: bi-directional motions in (a) are used to approximate forward bilateral motions in (b) and backward bilateral motions in (c).}
    \label{fig:intermediate motion approximation}
\end{figure*}

Fig.~\ref{fig:intermediate motion approximation} illustrates this motion approximation, in which each column represents a frame at a time instance and a dot corresponds to a pixel in the frame. In particular, in Fig.~\ref{fig:intermediate motion approximation}(a), an occluded pixel in $I_0$ is depicted by a green dot. To complement the inaccuracy of the bilateral motion at pixel $\bf x$ in $I_t$, we use two bi-directional motions $V_{0 \rightarrow 1}({\bf x})$ and $V_{1 \rightarrow 0}({\bf x})$, which are depicted by green and red lines, respectively. We approximate two forward bilateral motions $V^{fw}_{t \rightarrow 1}$ and $V^{fw}_{t \rightarrow 0}$ in Fig.~\ref{fig:intermediate motion approximation}(b) using $V_{0\rightarrow 1}$. Specifically, for pixel $\bf x$ in $I_t$, depicted by an orange dot, we approximate a motion vector $V^{fw}_{t \rightarrow 1}({\bf x})$ by scaling $V_{0 \rightarrow 1}({\bf x})$ with a factor $(1-t)$, assuming that the motion vector field is locally smooth. Since the bilateral motion estimation is based on the assumption that a motion trajectory between consecutive frames is linear, two approximate motions $V^{fw}_{t\rightarrow1}(\mathbf{x})$ and $V^{fw}_{t\rightarrow0}(\mathbf{x})$ should be symmetric with respect to $\bf x$ in $I_t$. Thus, we obtain an additional approximate vector $V^{fw}_{t\rightarrow0}(\mathbf{x})$ by reversing the direction of the vector $V^{fw}_{t\rightarrow1}(\mathbf{x})$. In other words, we approximate the forward bilateral motions by
\begin{align}
    \label{eq:app_forward_t1}
    V^{fw}_{t\rightarrow1}(\mathbf{x}) &= (1-t) \times V_{0\rightarrow1}(\mathbf{x}),\\
    \label{eq:app_forward_t0}
    V^{fw}_{t\rightarrow0}(\mathbf{x}) &=  (-t) \times V_{0\rightarrow1}(\mathbf{x}).
\end{align}
% Similarly, we approximate F0:5!0(x) from F1!0(x).
Similarly, we approximate the backward bilateral motions by
\begin{align}
    \label{eq:app_backward_t0}
    V^{bw}_{t\rightarrow0}(\mathbf{x}) &=  t \times V_{1\rightarrow0}(\mathbf{x}), \\
    \label{eq:app_backward_t1}
    V^{bw}_{t\rightarrow1}(\mathbf{x}) &= -(1-t) \times V_{1\rightarrow0}(\mathbf{x}),
\end{align}
as illustrated in Fig.~\ref{fig:intermediate motion approximation}(c). Note that Jiang \etal~\cite{jiang2018slomo} also used these equations \eqref{eq:app_forward_t1}$\sim$\eqref{eq:app_backward_t1}, but derived only two motion candidates: $V_{t \rightarrow 1}({\bf x})$ by combining \eqref{eq:app_forward_t1} and \eqref{eq:app_backward_t1} and $V_{t \rightarrow 0}({\bf x})$ by combining \eqref{eq:app_forward_t0} and \eqref{eq:app_backward_t0}. Thus, if an approximated motion in \eqref{eq:app_forward_t1}$\sim$\eqref{eq:app_backward_t1} is unreliable, the combined one is also degraded.
In contrast, we use all four candidates in \eqref{eq:app_forward_t1}$\sim$\eqref{eq:app_backward_t1} directly to choose reliable motions in Section~\ref{sec:synthesis}.

Fig.~\ref{fig:Intermediate candidates} shows that, whereas the bilateral motion estimation provides visual artifacts in (d), the motion approximation provides results without noticeable artifacts in (f). On the other hand, the bilateral motion estimation is more effective than the motion approximation in the cases of (e) and (g). Thus, the two schemes are complementary to each other.

\subsection{Frame synthesis}\label{sec:synthesis}

We interpolate an intermediate frame by combining the six intermediate candidates, which are warped by the warping layers in Fig.~\ref{fig:overview}. If we consider only color information, rich contextual information in the input frames may be lost during the synthesis~\cite{niklaus2018ctx, liu2019cyclicgen, ronneberger2015unet}, degrading the interpolation performance. Hence, as in~\cite{niklaus2018ctx, bao2018memc, bao2019dain}, we further exploit contextual information in the input frames, called context maps. Specifically, we extract the output of the conv1 layer of ResNet-18~\cite{he2016resnet} as a context map, which is done by the context extractor in Fig.~\ref{fig:overview}.

By warping the two input frames and the corresponding context maps, we obtain six pairs of a warped frame and its context map: two pairs are reconstructed using the bilateral motion estimation, and four pairs using the motion approximation. Fig.~\ref{fig:overview} shows these six pairs. Since these six warped pairs have different characteristics, they are used as complementary candidates of the intermediate frame. Recent video interpolation algorithms employ synthesis neural networks, which take warped frames as input and yield final interpolation results or residuals to refine pixel-wise blended results~\cite{bao2018memc, niklaus2018ctx, bao2019dain}. However, these synthesis networks may cause artifacts if motions are inaccurately estimated. To alleviate these artifacts, instead, we develop a dynamic filter network~\cite{jia2016dfn} that takes the aforementioned six pairs of candidates as input and outputs local blending filters, which are then used to process the warped frames to yield the intermediate frame. These local blending filters compensate for motion inaccuracies, by considering spatiotemporal neighboring pixels in the stack of warped frames. The frame synthesis layer performs this synthesis in Fig.~\ref{fig:overview}.

\subsubsection{Dynamic local blending filters:}

\begin{figure}[t]
\setlength{\belowcaptionskip}{-10pt}
  \centering
  \includegraphics[width=9cm]{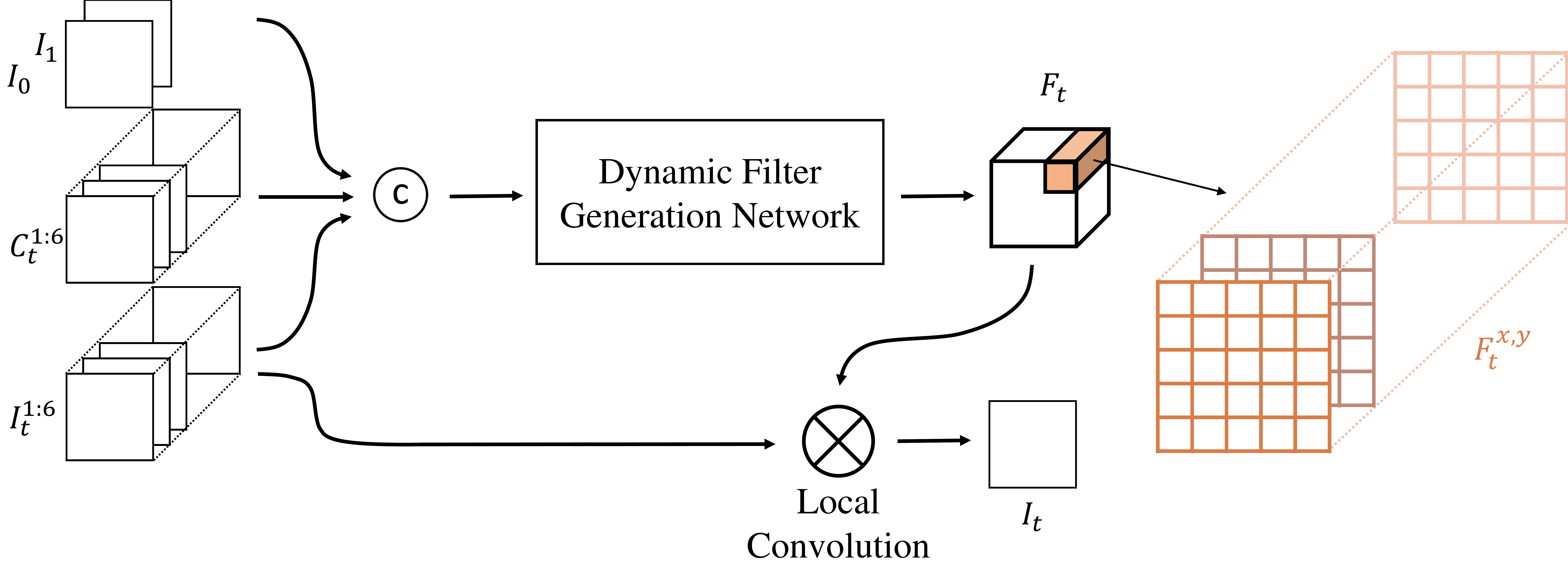}
  \caption
  {
    Frame synthesis using dynamic local blending filters.
  }
  \label{fig:Dynamic Filter}
\end{figure}

Fig.~\ref{fig:Dynamic Filter} shows the proposed synthesis network using dynamic local blending filters. The coefficients of the filters are learned from the images and contextual information through a dynamic blending filter network~\cite{jia2016dfn}. We employ the residual dense network~\cite{zhang2018rdn} as the backbone for the filter generation. In Fig.~\ref{fig:Dynamic Filter}, the generation network takes the input frames $I_0$ and $I_1$ and the intermediate candidates $\{I^{1:6}_{t}\}$ with the corresponding context maps $\{C^{1:6}_{t}\}$ as input. Then, for each pixel ${\bf x} = (x,y)$, we generate six blending filters to fuse the six intermediate candidates, given by
\begin{equation}
  F^{x,y}_t\in\mathbb{R}^{5\times5\times6}.
\end{equation}
For each $\bf x$, the sum of all coefficients in the six filters are normalized to 1.

Then, the intermediate frame is synthesized via the dynamic local convolution. More specifically, the intermediate frame is obtained by filtering the intermediate candidates, given by
\begin{equation}\label{eq:Local filtering}
  I_t(x,y)=\sum_{c=1}^{6}\:  \sum_{i=-2}^{2}  \sum_{j=-2}^{2}F^{x,y}_t(i,j,c)I^c_t(x+i,y+j).
\end{equation}

\subsection{Training}

The proposed algorithm includes two neural networks: the bilateral motion network and the dynamic filter generation network. We found that separate training of these two networks is more efficient than the end-to-end training in training time and memory space. Thus, we first train the bilateral motion network. Then, after fixing it, we train the dynamic filter generation network.

\subsubsection{Bilateral motion network:}

To train the proposed bilateral motion network, we define the bilateral loss $\mathcal{L}_b$ as
\begin{equation}\label{eq:Loss for bilateral motion network}
  \mathcal{L}_b = \mathcal{L}_p + \mathcal{L}_s
\end{equation}
where $\mathcal{L}_p$ and $\mathcal{L}_s$ are the photometric loss~\cite{Yu2016UnOpt, Ren2017Unopt} and the smoothness loss~\cite{liu2017dvf}.

For the photometric loss, we compute the sum of differences between a ground-truth frame $I^l_t$ and two warped frames $I^l_{0\rightarrow t}$ and $I^l_{1\rightarrow t}$ using the bilateral motion fields $V^l_{t \rightarrow 0}$ and $V^l_{t \rightarrow 1}$, respectively, at all pyramid levels,
\begin{equation}
    \label{eq:photometric loss}
    \mathcal{L}_p = \sum_{l=1}^{L} \alpha_l \bigg[ \sum_{\mathbf{x}} \rho(I^l_{0\rightarrow t}(\mathbf{x})-I^l_{t}(\mathbf{x}))
  +\rho(I^l_{1\rightarrow t} (\mathbf{x})-I^l_{t}(\mathbf{x}))\bigg]
\end{equation}
where $\rho(x)=\sqrt{x^2+\epsilon^2}$ is the Charbonnier function~\cite{Char1994loss}.
The parameters $\alpha_l$ and $\epsilon$ are set to $0.01\times2^l$ and $10^{-6}$, respectively. Also, we compute the smoothness loss to constrain neighboring pixels to have similar motions, given by
\begin{equation}
    \label{eq:smoothness loss}
    \mathcal{L}_s = \| \nabla V_{t\rightarrow 0} \|_1 + \| \nabla V_{t\rightarrow 1} \|_1.
\end{equation}

We use the Adam optimizer~\cite{kingma2014adam} with a learning rate of $\eta = 10^{-4}$ and shrink it via $\eta\leftarrow 0.5\eta$ at every 0.5M iterations. We use a batch size of 4 for 2.5M iterations and augment the training dataset by randomly cropping $256\times256$ patches with random flipping and rotations.

\subsubsection{Dynamic filter generation network:}
We define the dynamic filter loss $\mathcal{L}_d$ as the Charbonnier loss between $I_t$ and its synthesized version $\hat{I}_t$, given by
\begin{equation}
    \label{eq:reconstruction loss}
    \mathcal{L}_d = \sum_{\mathbf{x}}\rho(\hat{I}_t(\mathbf{x})-I_t(\mathbf{x})).
\end{equation}

Similarly to the bilateral motion network, we use the Adam optimizer with $\eta = 10^{-4}$ and shrink it via $\eta\leftarrow 0.5\eta$ at 0.5M, 0.75M, and 1M iterations. We use a batch size of 4 for 1.25M iterations. Also, we use the same augmentation technique as that for the bilateral motion network.

\subsubsection{Datasets:}

We use the Vimeo90K dataset~\cite{xue2019toflow} to train the proposed networks.
The training set in Vimeo90K is composed of 51,312 triplets with a resolution of $448\times256$. We train the bilateral motion network with $t = 0.5 $ at the first 1M iterations and then with $t\in\{0,0.5,1\}$ for fine-tuning. Next, we train the dynamic filter generation network with $t = 0.5$. However, notice that both networks are capable of handling any $t \in (0,1)$ using the bilateral cost volume in (\ref{eq:bilateral_cost_volume}).

\section{Experimental Results}

We evaluate the performances of the proposed video interpolation algorithm on the Middlebury~\cite{baker2011database}, Vimeo90K~\cite{xue2019toflow}, UCF101~\cite{soomro2012ucf}, and Adobe240-fps~\cite{Su2017adobe} datasets.
We compare the proposed algorithm with state-of-the-art algorithms. Then, we conduct ablation studies to analyze the contributions of the proposed bilateral motion network and dynamic filter generation network.

\subsection{Datasets}

%\subsubsection{Middlebury:}
{\bf Middlebury:}
The Middlebury benchmark~\cite{baker2011database}, the most commonly used benchmark for video interpolation, provides two sets: Other and Evaluation.
`Other' contains the ground-truth for fine-tuning, while `Evaluation' provides two frames selected from each of 8 sequences for evaluation.\\

%\subsubsection{Vimeo90K:}
{\noindent \bf Vimeo90K:}
The test set in Vimeo90K~\cite{xue2019toflow} contains 3,782 triplets of spatial resolution $256\times448$. It is not used to train the model.\\

%\subsubsection{UCF101:}
{\noindent \bf UCF101:}
The UCF101 dataset~\cite{soomro2012ucf} contains human action videos of resolution $256\times256$. Liu {\it et al.}~\cite{liu2017dvf} constructed the test set by selecting 379 triplets.\\

%\subsubsection{Adobe240-fps:}
{\noindent \bf Adobe240-fps:}
Adobe240-fps~\cite{Su2017adobe} consists of high frame-rate videos. To assess the interpolation performance, we selected a test set of 254 sequences, each of which consists of nine frames.

\subsection{Comparison with the state-of-the-arts}

We assess the interpolation performances of the proposed algorithm in comparison with the conventional video interpolation algorithms:
MIND ~\cite{long2016learning}, DVF~\cite{liu2017dvf}, SpyNet~\cite{ranjan2017spynet}, SepConv~\cite{niklaus2017sepconv}, CtxSyn~\cite{niklaus2018ctx},
ToFlow \cite{xue2019toflow}, SuperSloMo~\cite{jiang2018slomo}, MEMC-Net~\cite{bao2018memc}, CyclicGen~\cite{liu2019cyclicgen}, and DAIN~\cite{bao2019dain}.
For SpyNet, we generated intermediate frames using the Baker \etal's algorithm~\cite{baker2011database}.

\begin{table*}[t]
    \caption
    {
        Quantitative comparisons on the Middlebury Evaluation set. For each metric, the numbers in \textcolor{red}{\bf red} and \textcolor{blue}{\underline{blue}} denote the best and the second best results, respectively.
    }
    \centering
    {\tiny
    \begin{tabular}{L{1.8cm}C{0.45cm}C{0.45cm}C{0.45cm}C{0.45cm}C{0.45cm}C{0.45cm}C{0.45cm}C{0.45cm}C{0.45cm}C{0.45cm}C{0.45cm}C{0.45cm}C{0.45cm}C{0.45cm}C{0.45cm}C{0.45cm}|C{0.45cm}C{0.45cm}}
    \toprule
    \multirow{2}[2]{*}{} & \multicolumn{2}{c}{Mequon} & \multicolumn{2}{c}{Schefflera} & \multicolumn{2}{c}{Urban} & \multicolumn{2}{c}{Teddy} & \multicolumn{2}{c}{Backyard} & \multicolumn{2}{c}{Basketball} & \multicolumn{2}{c}{Dumptruck} & \multicolumn{2}{c}{Evergreen} & \multicolumn{2}{c}{Average}\\[-0.1em]
    \cmidrule(lr){2-3} \cmidrule(lr){4-5} \cmidrule(lr){6-7} \cmidrule(lr){8-9} \cmidrule(lr){10-11} \cmidrule(lr){12-13} \cmidrule(lr){14-15} \cmidrule(lr){16-17} \cmidrule(lr){18-19}
    & IE & NIE & IE & NIE & IE & NIE & IE & NIE & IE & NIE & IE & NIE & IE & NIE & IE & NIE & IE & NIE \\[-0.1em]
    \midrule

    SepConv-$L_1$\cite{niklaus2017sepconv} &2.52& \textcolor{blue}{\underline{0.54}}&3.56&0.67&4.17&1.07&5.41&1.03&10.2&0.99&5.47&0.96&6.88&0.68&6.63&0.70&5.61&0.83\\
    ToFlow\cite{xue2019toflow} &2.54 &0.55& 3.70 &0.72 &3.43 &0.92 &5.05 &0.96 &9.84 &0.97 &5.34 &0.98 &6.88 &0.72 &7.14 & 0.90 &5.49 &0.84\\
    SuperSlomo\cite{jiang2018slomo} &2.51 &0.59 &3.66 &0.72 & \textcolor{red}{\bf 2.91} &\textcolor{blue}{\underline{0.74}} &5.05 &0.98 &9.56 &0.94 &5.37 &0.96 &6.69 &0.60 & 6.73& 0.69 &5.31 &0.78\\
    CtxSyn\cite{niklaus2018ctx} &2.24 &\textcolor{red}{\bf 0.50} & \textcolor{red}{\bf 2.96} & \textcolor{red}{\bf 0.55} &4.32 &1.42 &\textcolor{red}{\bf 4.21} &0.87 &9.59 &0.95 &5.22 &0.94 &7.02 &0.68 &6.66 &0.67 &5.28 &0.82\\
    MEMC-Net*\cite{bao2018memc} &2.47 &0.60 &3.49 &0.65 &4.63 &1.42 &4.94 &0.88 &8.91 &0.93 &\textcolor{blue}{\underline{4.70}} &0.86 &6.46 &0.66 &6.35 & \textcolor{blue}{\underline{0.64}} &5.24 &0.83\\
    DAIN\cite{bao2019dain} & \textcolor{blue}{\underline{2.38}} &0.58 &3.28 &0.60 &3.32 &\textcolor{red}{\bf 0.69} &4.65 &\textcolor{blue}{\underline{0.86}} &\textcolor{blue}{\underline{7.88}} &\textcolor{blue}{\underline{0.87}} &4.73 &\textcolor{blue}{\underline{0.85}} &\textcolor{blue}{\underline{6.36}} &\textcolor{blue}{\underline{0.59}} &\textcolor{blue}{\underline{6.25}} &0.66 &\textcolor{blue}{\underline{4.86}} &\textcolor{blue}{\underline{0.71}}\\
    BMBC (Ours) & \textcolor{red}{\bf 2.30} & 0.57 & \textcolor{blue}{\underline{3.07}} & \textcolor{blue}{\underline{0.58}} & \textcolor{blue}{\underline{3.17}} & 0.77 & \textcolor{blue}{\underline{4.24}} & \textcolor{red}{\bf 0.84} & \textcolor{red}{\bf 7.79} & \textcolor{red}{\bf 0.85} & \textcolor{red}{\bf 4.08} & \textcolor{red}{\bf 0.82} & \textcolor{red}{\bf 5.63} & \textcolor{red}{\bf 0.58} & \textcolor{red}{\bf 5.55} & \textcolor{red}{\bf 0.56} & \textcolor{red}{\bf 4.48} & \textcolor{red}{\bf 0.70}\\[-0.1em]
    \bottomrule\\[-2.5em]
    \end{tabular}
    }
    \label{table:Evaluation on Middlebury EVAL set}
\end{table*}

\begin{figure*}[t]
    \setlength{\belowcaptionskip}{-10pt}
    \centering
	
    \subfloat {\includegraphics[width=1.45cm]{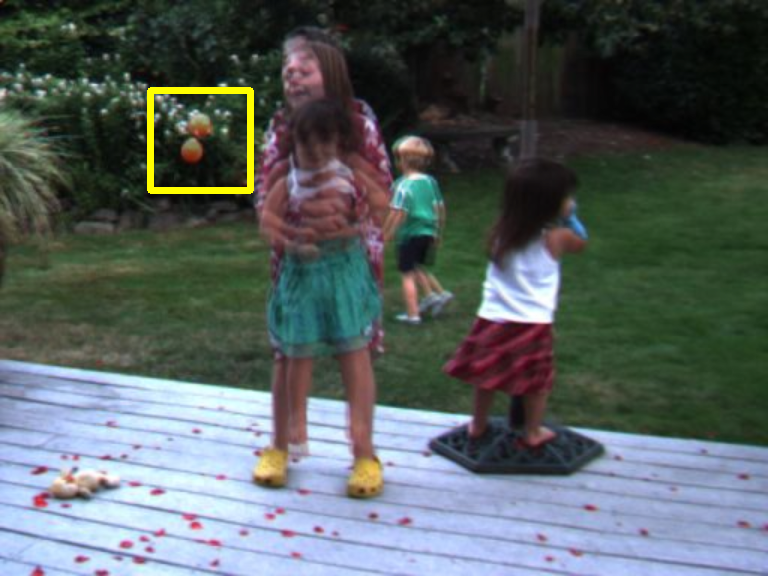}}\!
	\subfloat {\includegraphics[width=1.45cm]{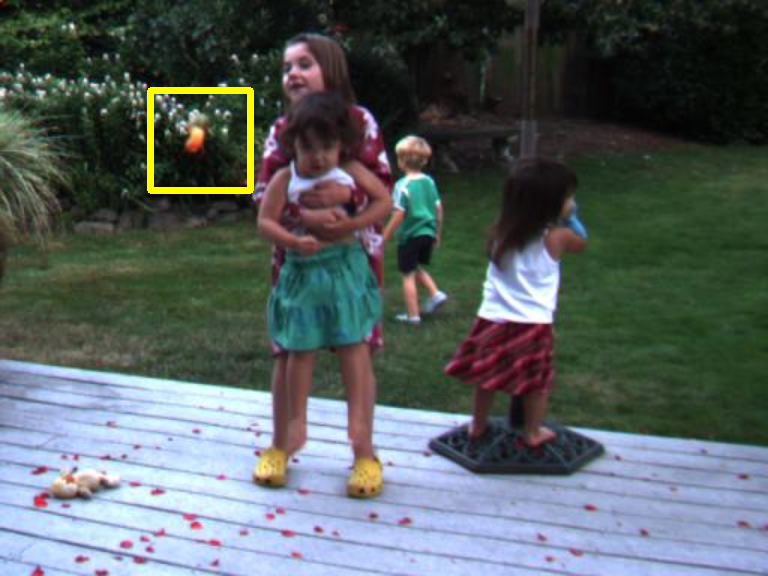}}\!
	\subfloat {\includegraphics[width=1.45cm]{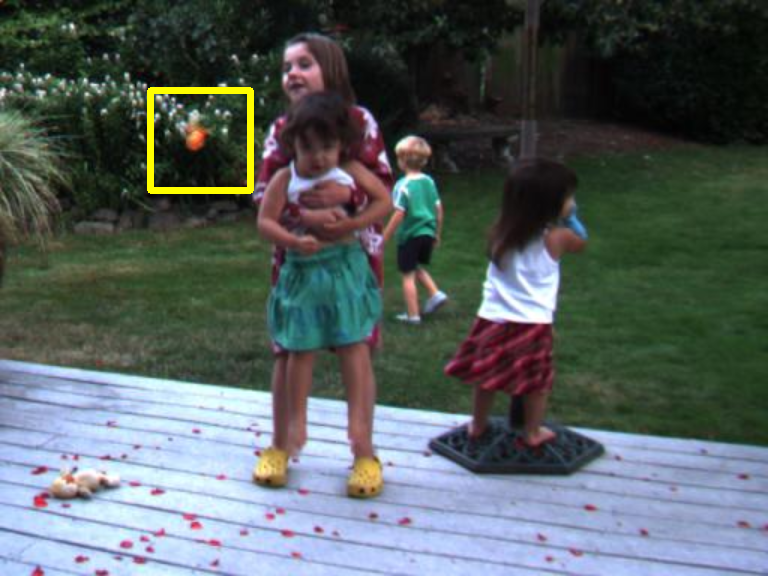}}\!
	\subfloat {\includegraphics[width=1.45cm]{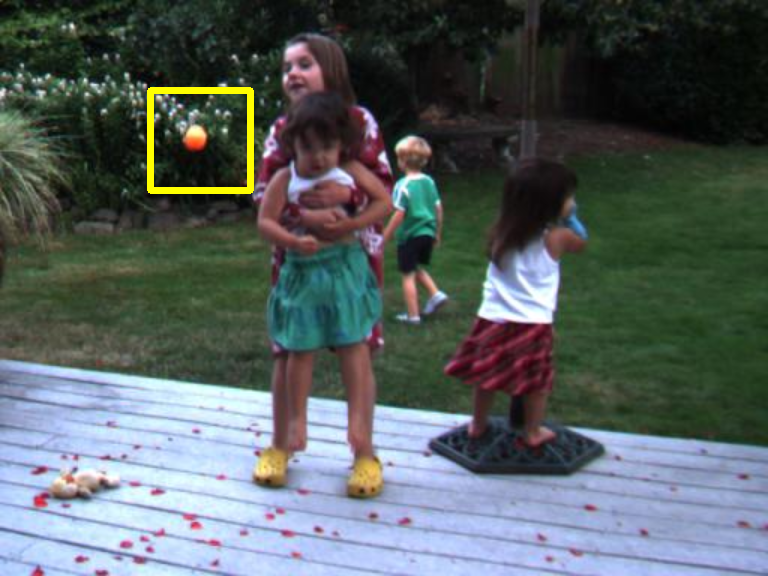}}\!
	\subfloat {\includegraphics[width=1.45cm]{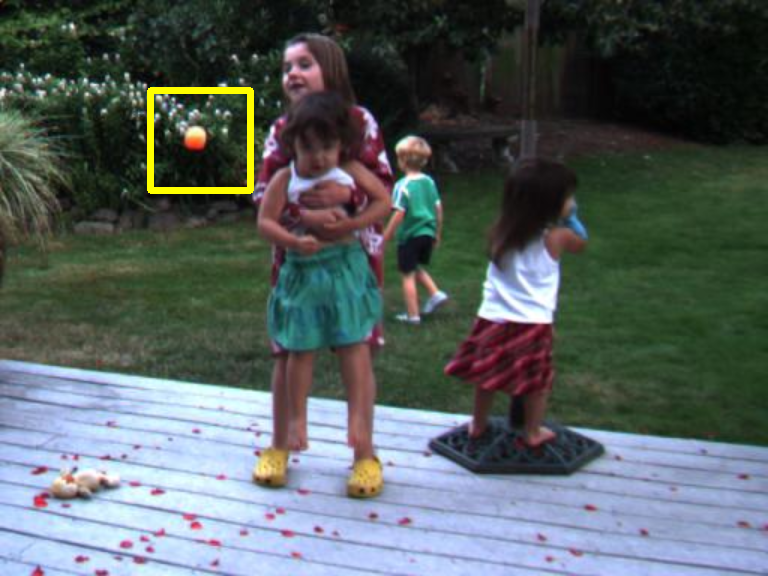}}\!
	\subfloat {\includegraphics[width=1.45cm]{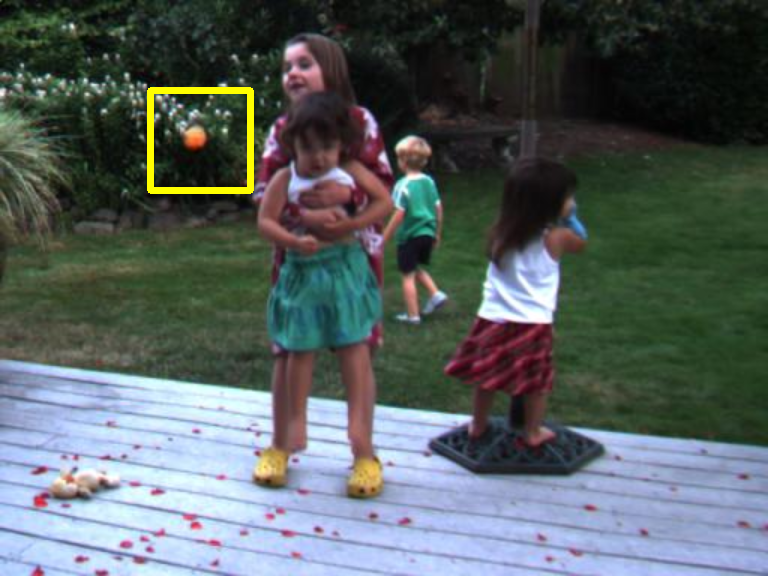}}\!
	\subfloat {\includegraphics[width=1.45cm]{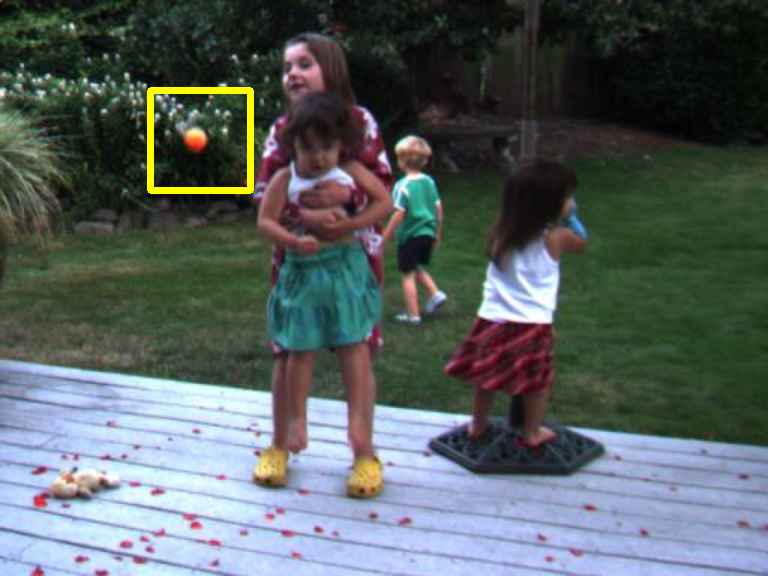}}\!
	\subfloat {\includegraphics[width=1.45cm]{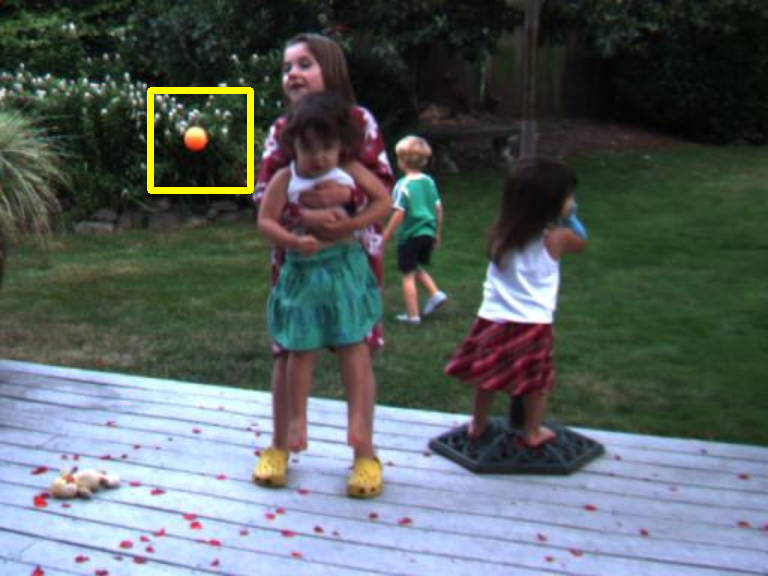}}\\[-2.5ex]

	%second row
    \setcounter{subfigure}{0}
    \subfloat {\includegraphics[width=1.45cm]{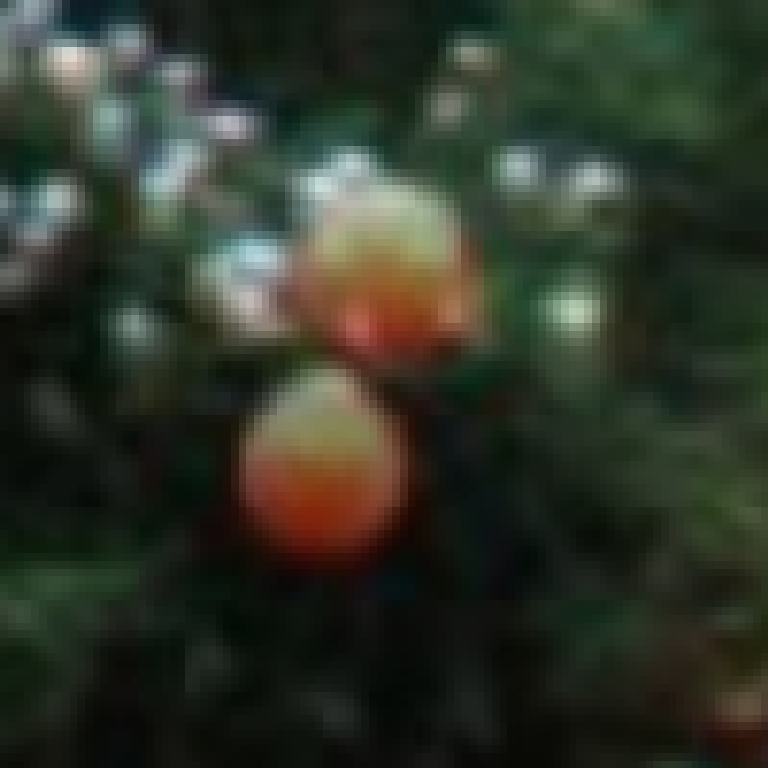}}\!
	\subfloat {\includegraphics[width=1.45cm]{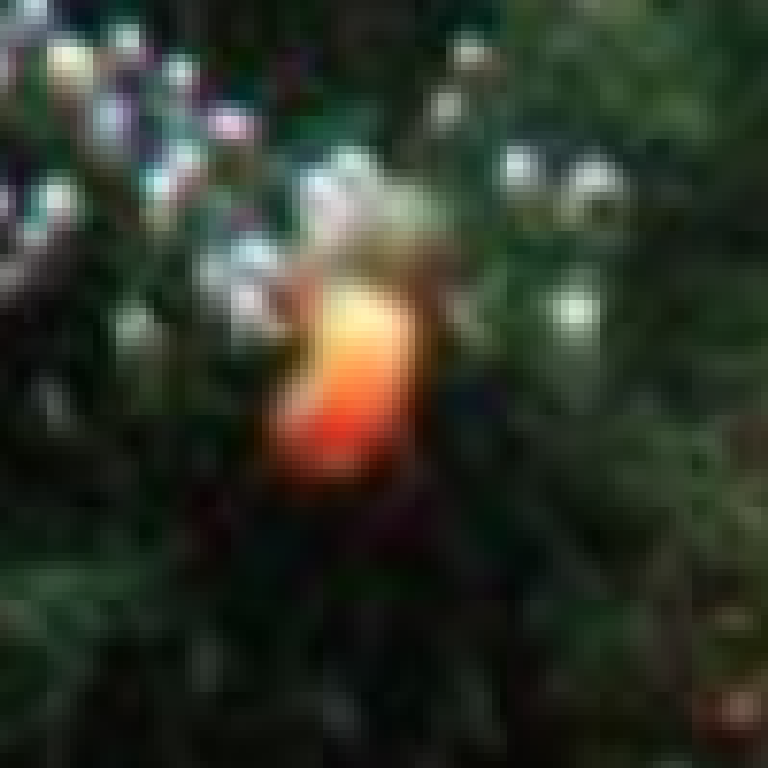}}\!
	\subfloat {\includegraphics[width=1.45cm]{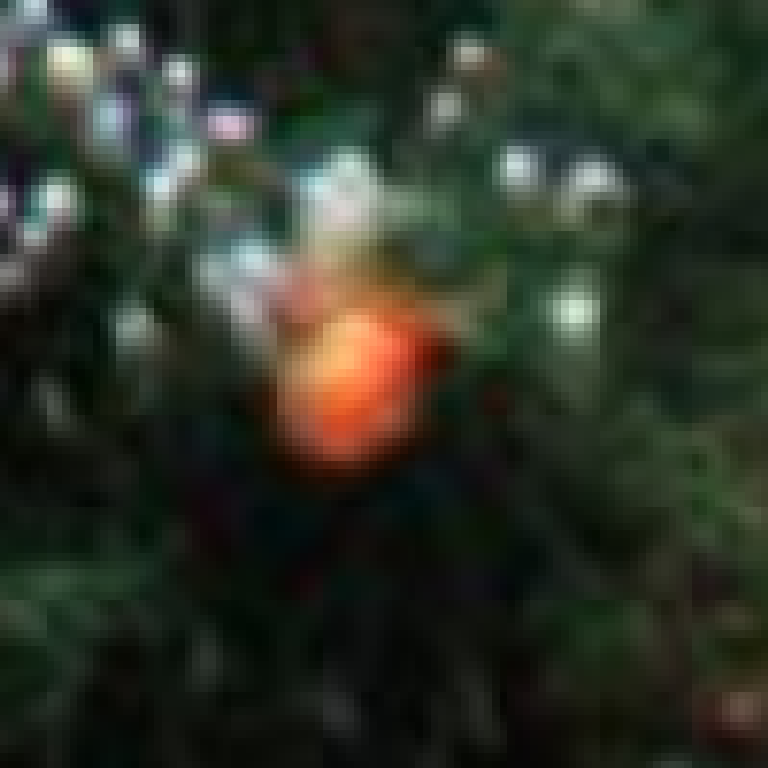}}\!
	\subfloat {\includegraphics[width=1.45cm]{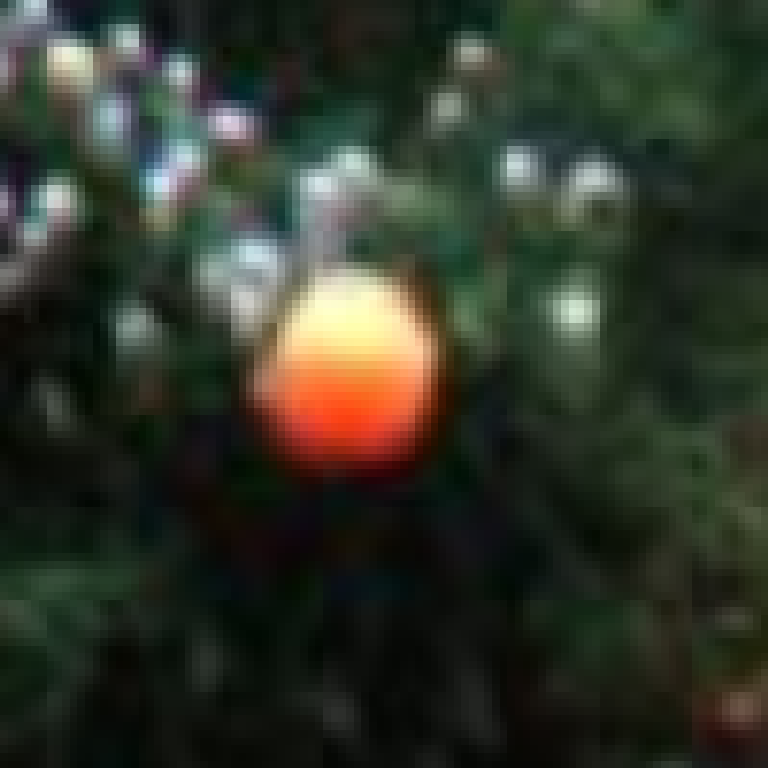}}\!
	\subfloat {\includegraphics[width=1.45cm]{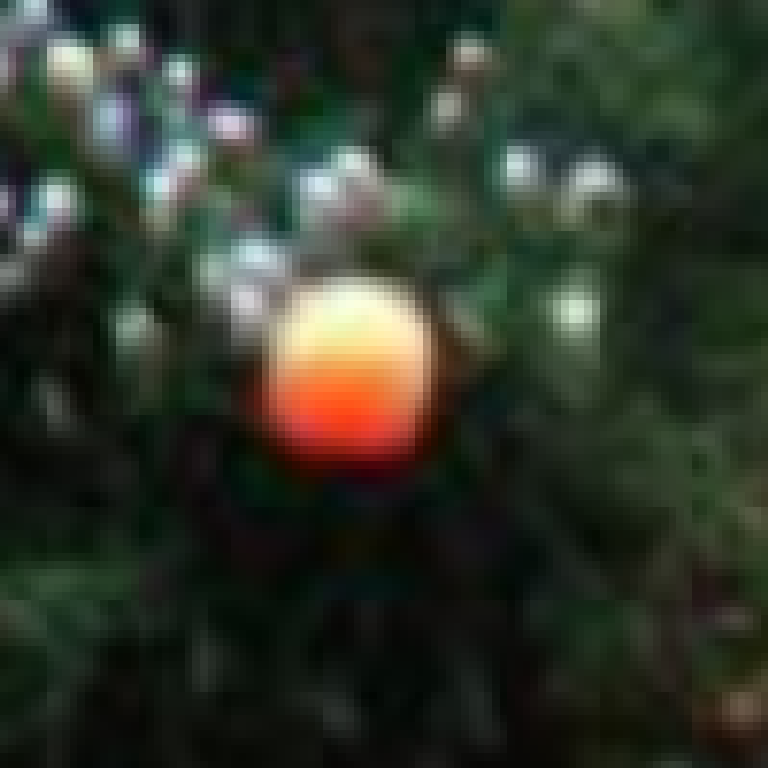}}\!
	\subfloat {\includegraphics[width=1.45cm]{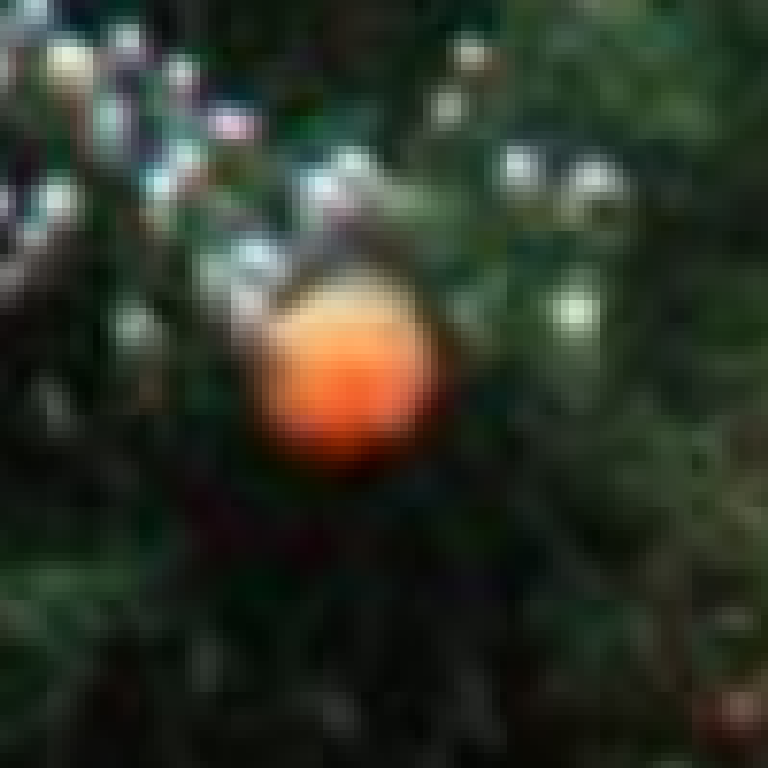}}\!
	\subfloat {\includegraphics[width=1.45cm]{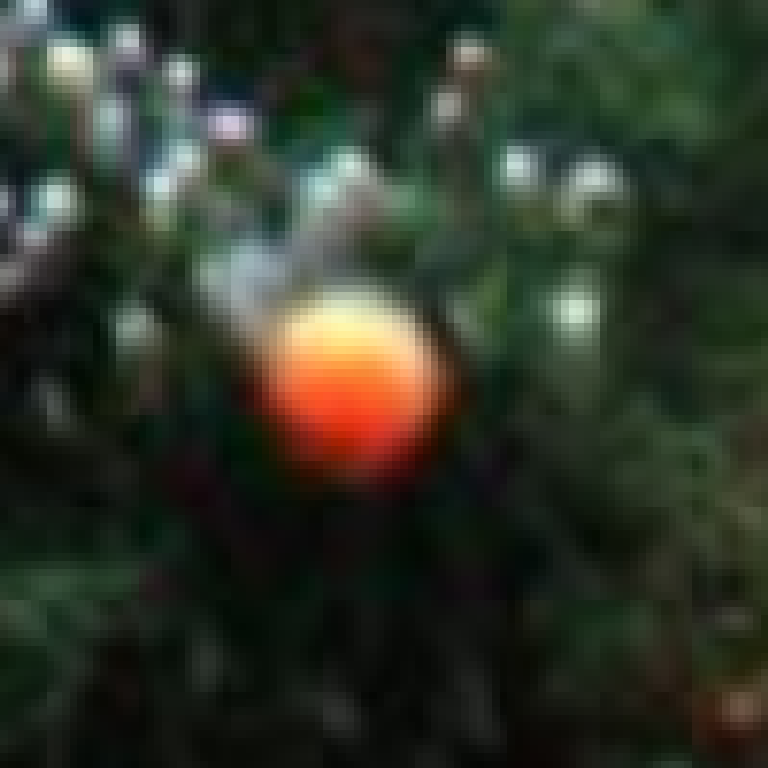}}\!
	\subfloat {\includegraphics[width=1.45cm]{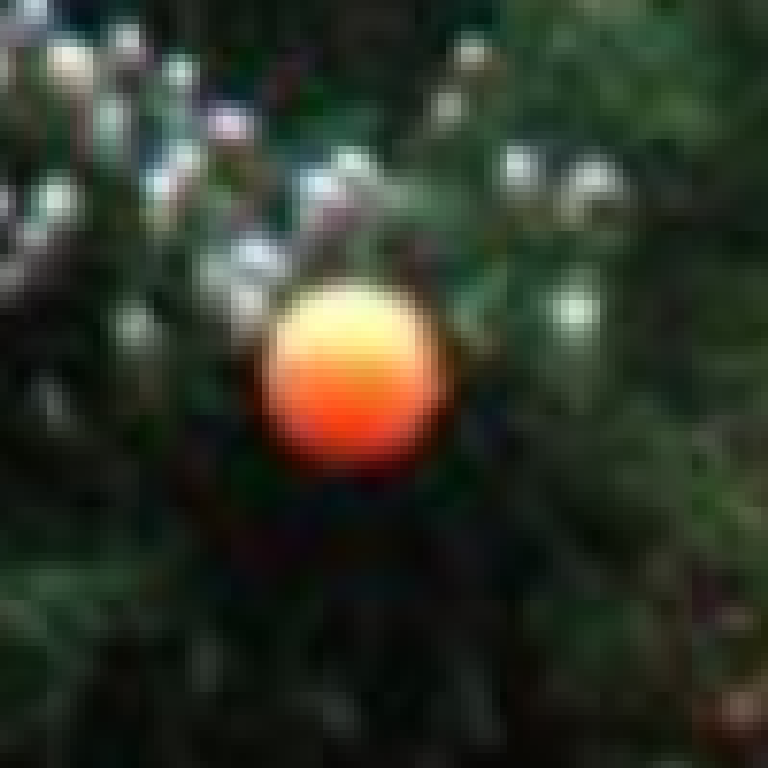}}\\[-2.5ex]

    \subfloat {\includegraphics[width=1.45cm]{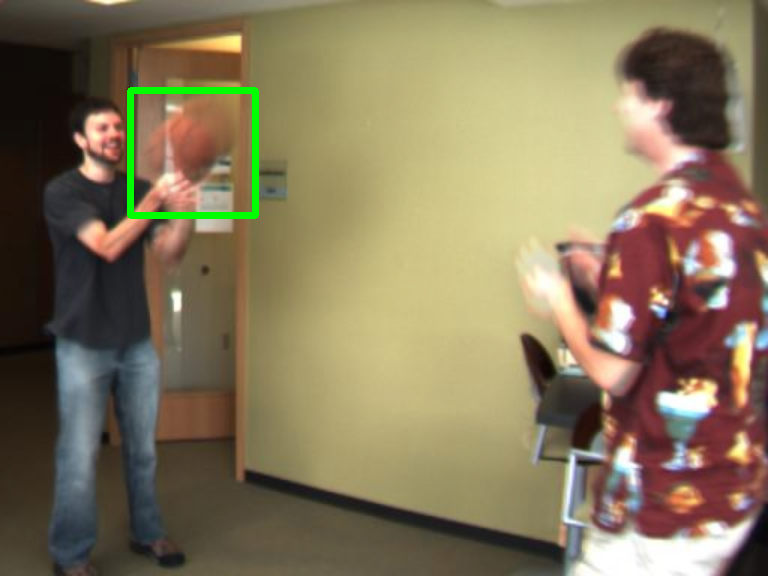}}\!
	\subfloat {\includegraphics[width=1.45cm]{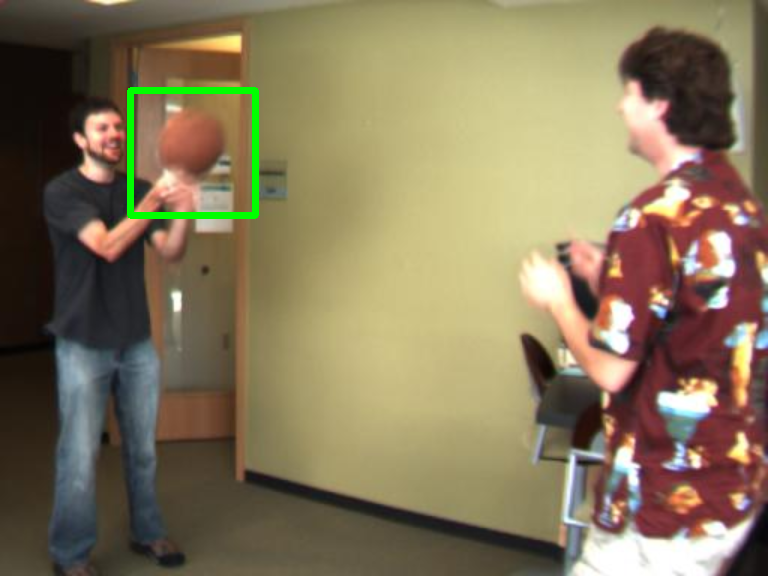}}\!
	\subfloat {\includegraphics[width=1.45cm]{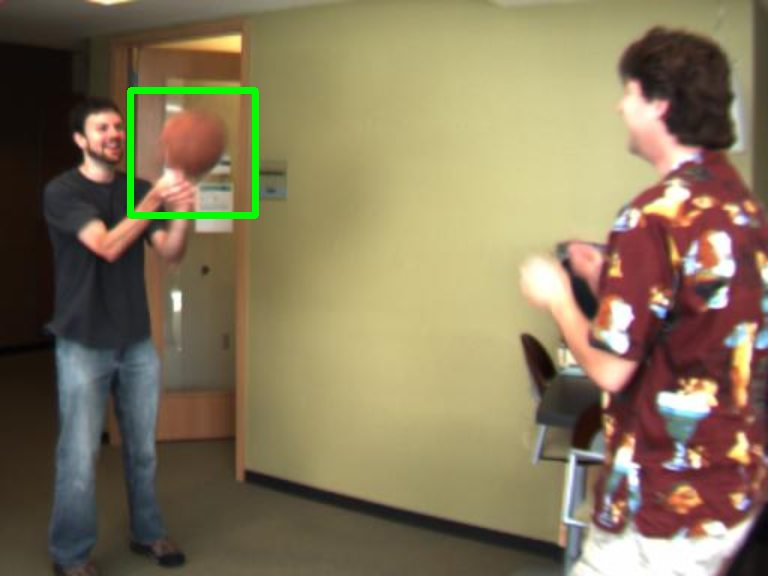}}\!
	\subfloat {\includegraphics[width=1.45cm]{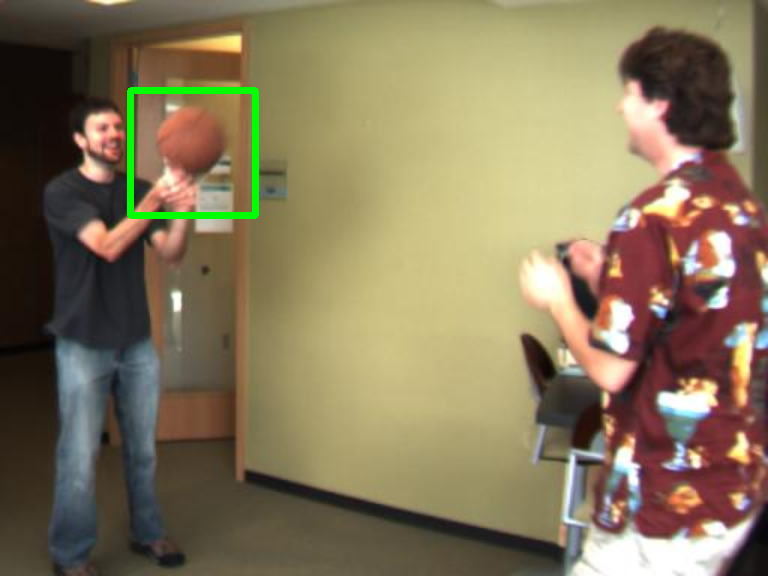}}\!
	\subfloat {\includegraphics[width=1.45cm]{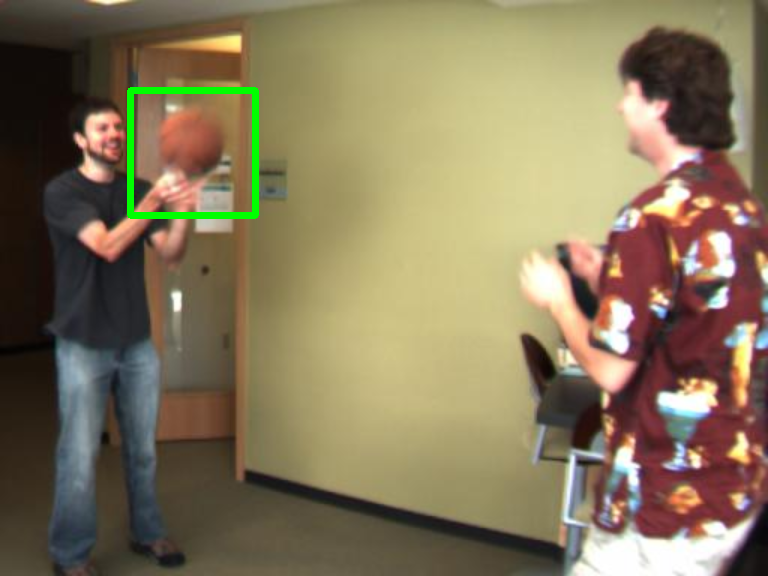}}\!
	\subfloat {\includegraphics[width=1.45cm]{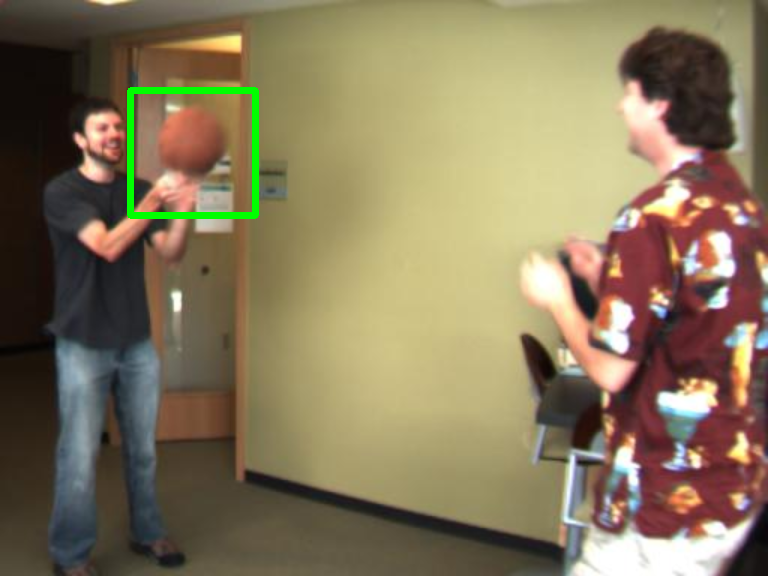}}\!
	\subfloat {\includegraphics[width=1.45cm]{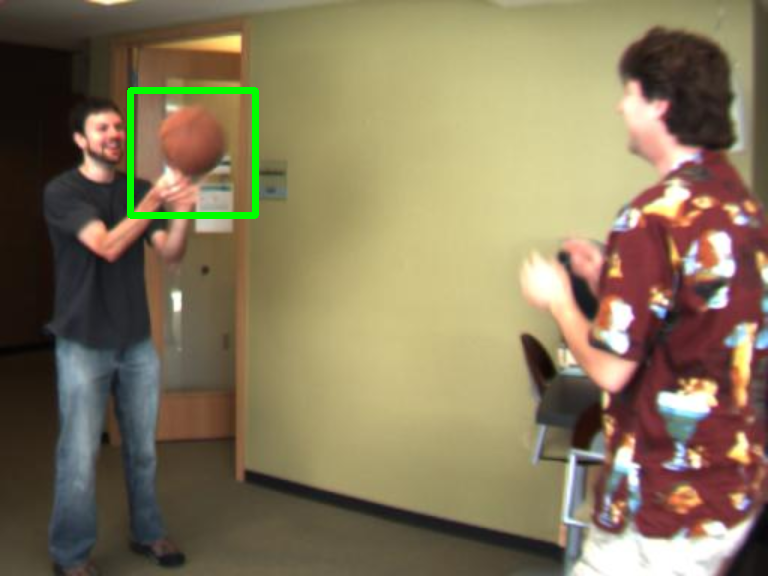}}\!
	\subfloat {\includegraphics[width=1.45cm]{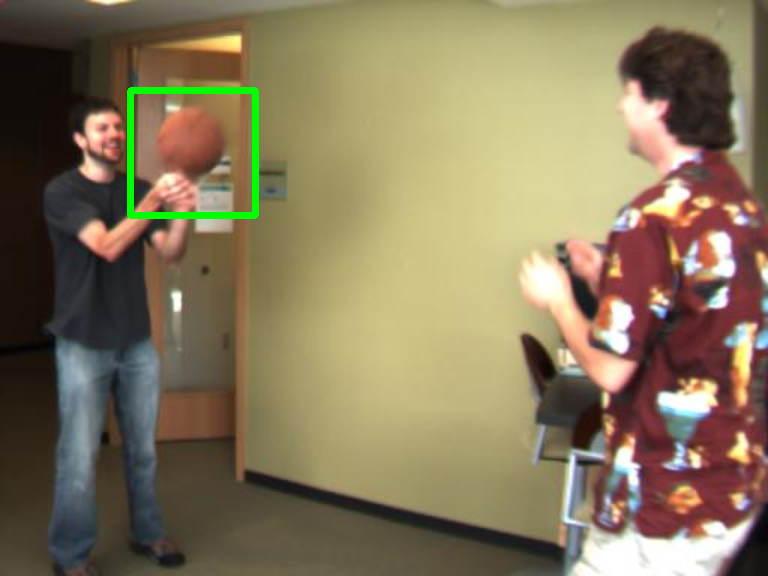}}\\[-2.5ex]

    \setcounter{subfigure}{0}
    \subfloat [] {\includegraphics[width=1.45cm]{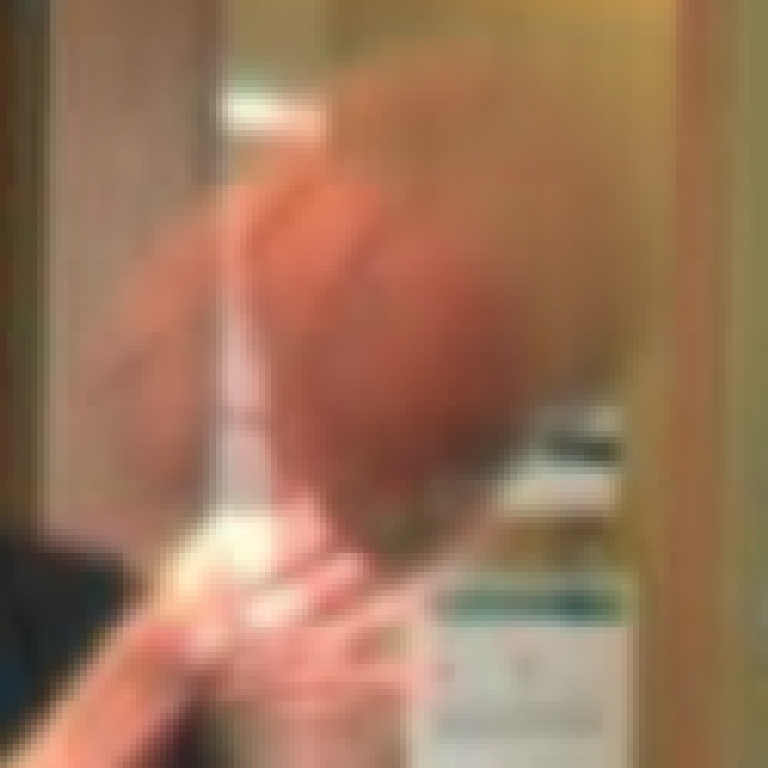}}\!
	\subfloat [] {\includegraphics[width=1.45cm]{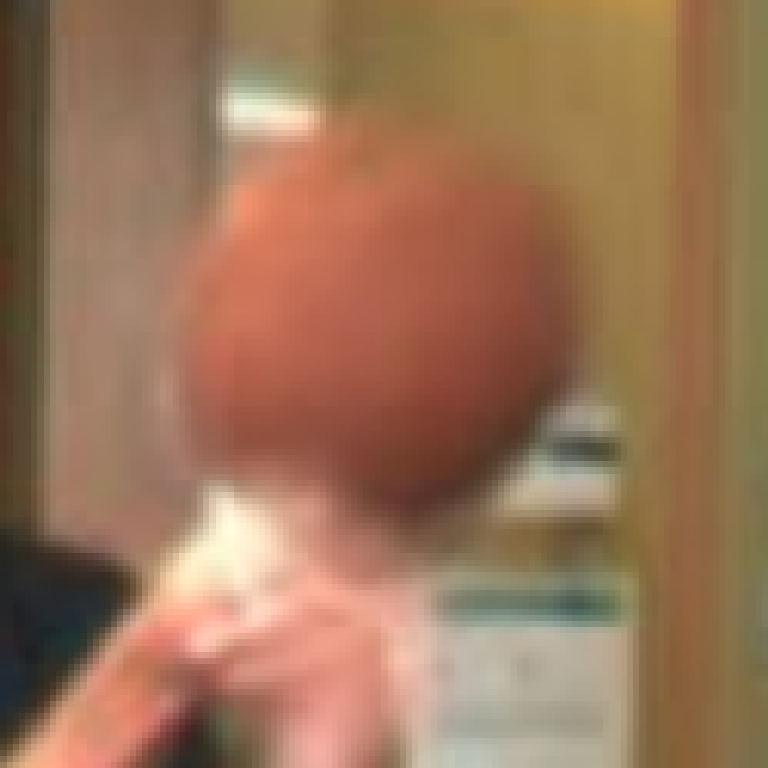}}\!
	\subfloat [] {\includegraphics[width=1.45cm]{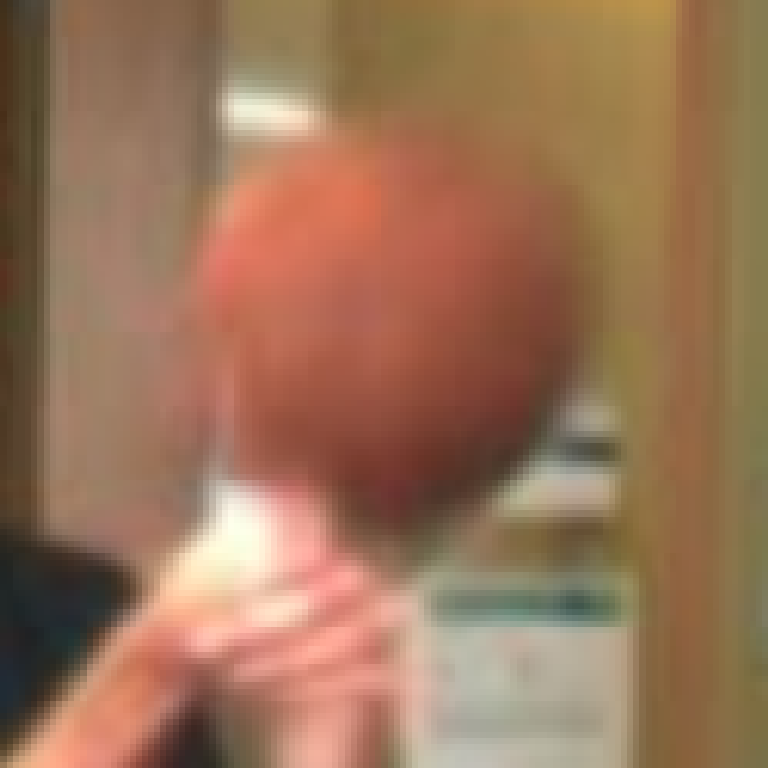}}\!
	\subfloat [] {\includegraphics[width=1.45cm]{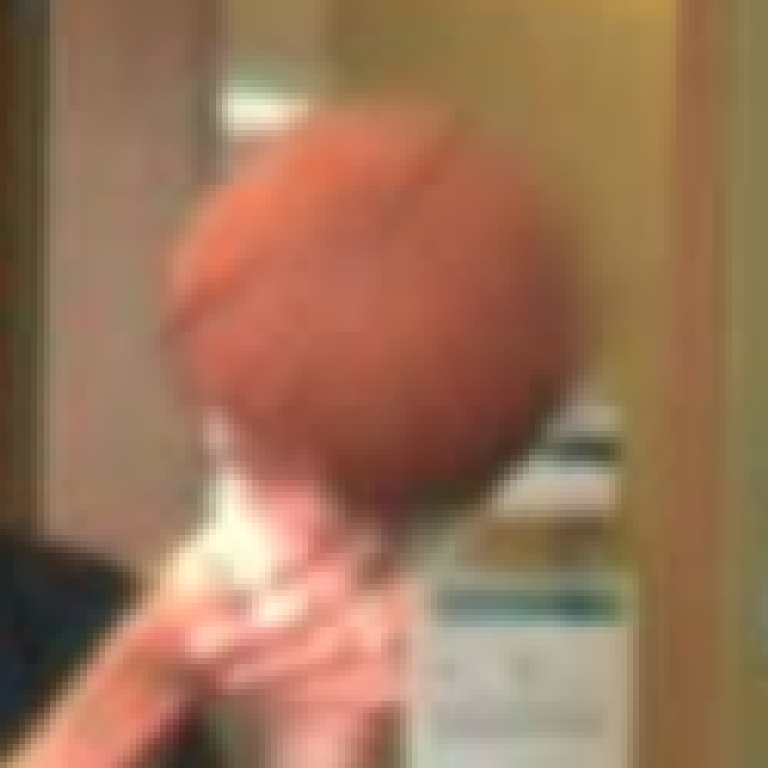}}\!
	\subfloat [] {\includegraphics[width=1.45cm]{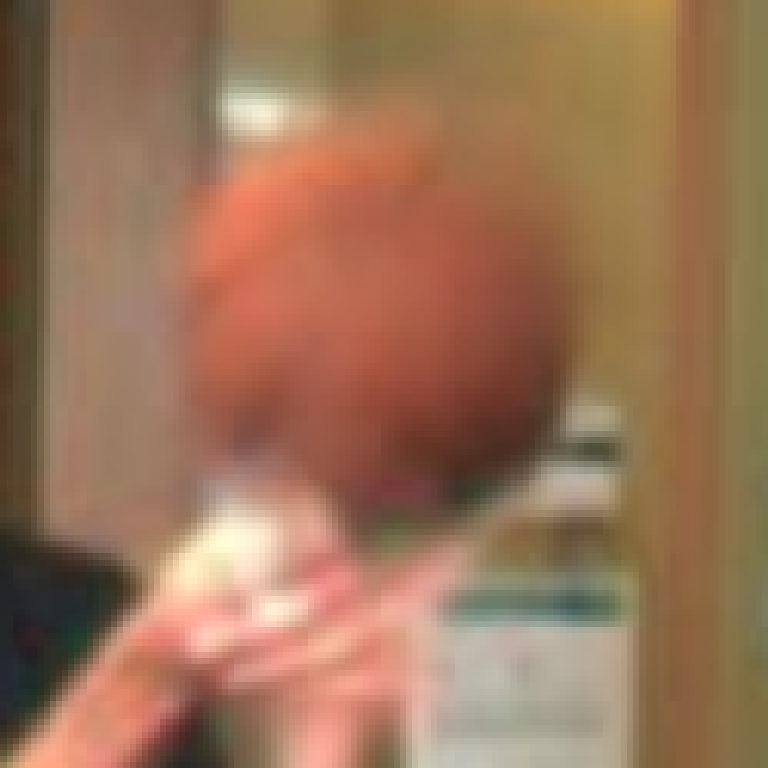}}\!
	\subfloat [] {\includegraphics[width=1.45cm]{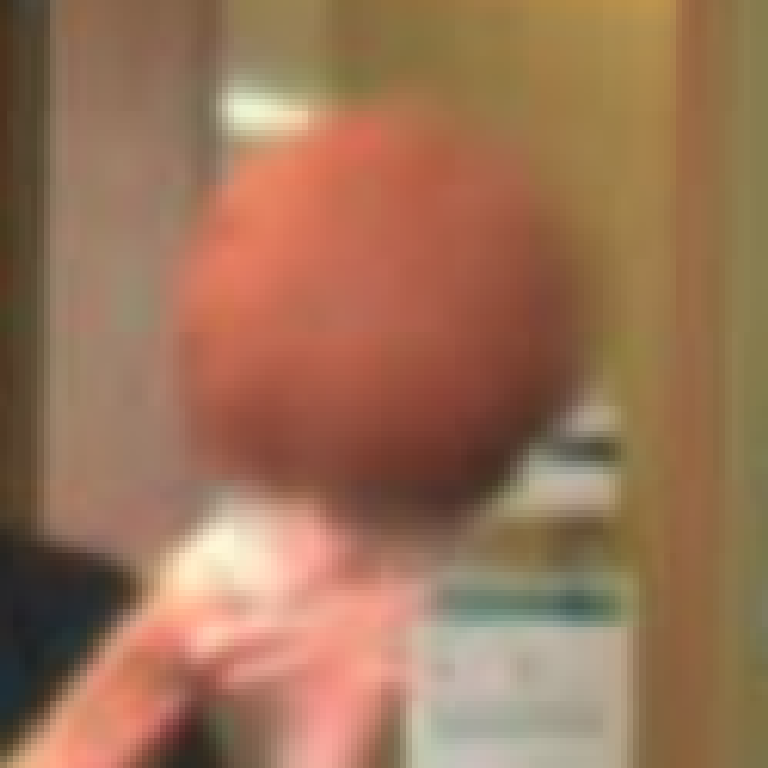}}\!
	\subfloat [] {\includegraphics[width=1.45cm]{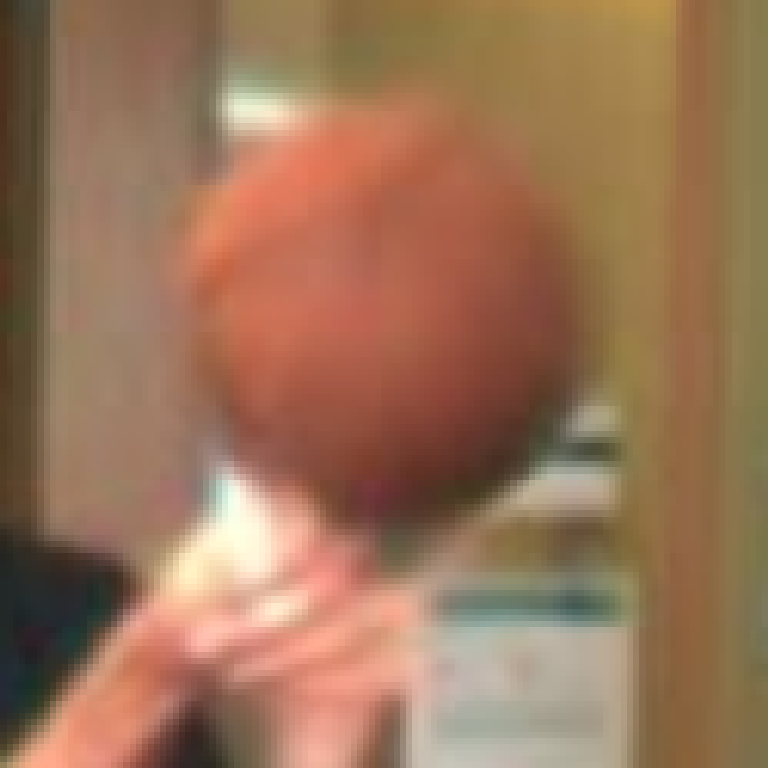}}\!
	\subfloat [] {\includegraphics[width=1.45cm]{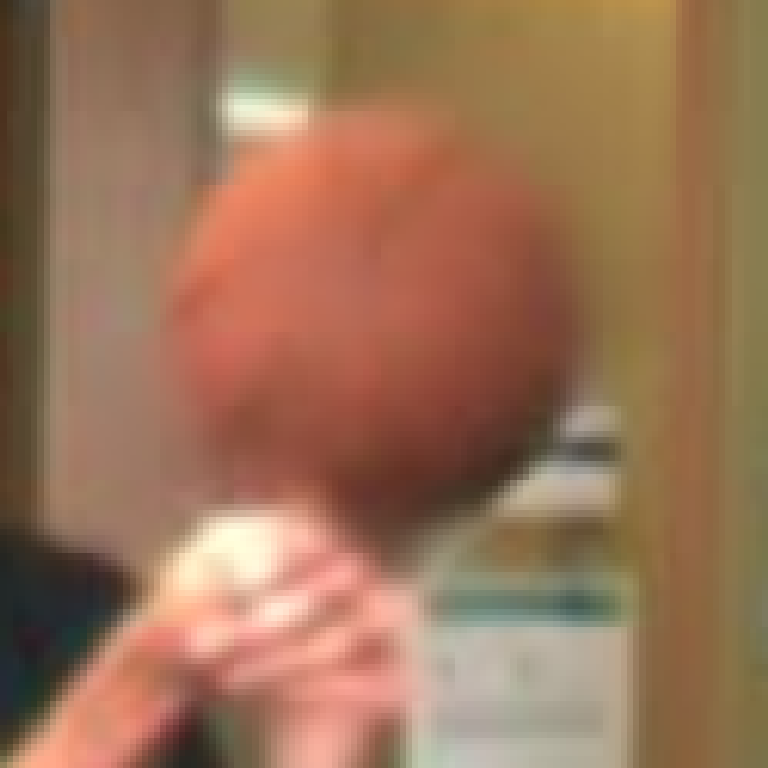}}\\[-0.5em]
	\caption
	{
        Visual comparison on the Middlebury Evaluation set. (a) Input, (b) SepConv-$L_1$~\cite{niklaus2017sepconv}, (c) ToFlow~\cite{xue2019toflow}, (d) SuperSlomo~\cite{jiang2018slomo}, (e) CtxSyn~\cite{niklaus2018ctx}, (f) MEMC-Net*~\cite{bao2018memc}, (g)~DAIN \cite{bao2019dain}, and (h) BMBC (Ours).
	}
	\label{fig:Eval on middlebury}
\end{figure*}

Table~\ref{table:Evaluation on Middlebury EVAL set} shows the comparisons on the Middlebury Evaluation set~\cite{baker2011database}, which are also available on the Middlebury website. We compare the average interpolation error (IE) and normalized interpolation error (NIE). A lower IE or NIE indicates better performance. The proposed algorithm outperforms all the state-of-the-art algorithms in terms of average IE and NIE scores. Fig.~\ref{fig:Eval on middlebury} visually compares interpolation results. SepConv-$L_1$~\cite{niklaus2017sepconv}, ToFlow~\cite{xue2019toflow} SuperSlomo~\cite{jiang2018slomo}, CtxSyn~\cite{niklaus2018ctx}, MEMC-Net~\cite{bao2018memc}, and DAIN~\cite{bao2019dain} yield blurring artifacts around the balls, losing texture details. On the contrary, the proposed algorithm  reconstructs the clear shapes of the balls, preserving the details faithfully.

\begin{table*}[t]
    \caption
    {
        Quantitative comparisons on the UCF101 and Vimeo90K datasets.
    }
    \centering
    {\scriptsize
    \begin{tabular}{L{2.5cm}C{1.5cm}C{2cm}C{1cm}C{1cm}C{1cm}C{1cm}}
    \toprule
    \multirow{2}[2]{*}{} & \multirow{2}[2]{*}{\makecell{Runtime \\ (seconds)}} & \multirow{2}[2]{*}{\makecell{\#Parameters \\ (million)}} & \multicolumn{2}{c}{UCF101\cite{liu2017dvf}} & \multicolumn{2}{c}{Vimeo90K~\cite{xue2019toflow}}\\[-0.1em]
    \cmidrule(lr){4-5} \cmidrule(lr){6-7}
    & & & PSNR & SSIM & PSNR & SSIM\\[-0.1em]
    \midrule

    SpyNet\cite{ranjan2017spynet}&0.11  &1.20 &33.67 &0.9633 &31.95 &0.9601\\
    MIND\cite{long2016learning}&0.01 &7.60 &33.93 &0.9661 &33.50 &0.9429\\
    DVF\cite{liu2017dvf}&0.47  &1.60 &34.12 &0.9631 &31.54 &0.9462\\
    ToFlow\cite{xue2019toflow}&0.43 &1.07 &34.58 &0.9667 &33.73 &0.9682\\
    SepConv-$L_f$\cite{niklaus2017sepconv}&0.20 &21.6&34.69 &0.9655 &33.45 &0.9674\\
    SepConv-$L_1$\cite{niklaus2017sepconv}&0.20 &21.6 &34.78 &0.9669 &33.79 &0.9702\\
    MEMC-Net\cite{bao2018memc}&0.12 &70.3 & 34.96 &0.9682 &34.29 &0.9739 \\
    CyclicGen\cite{liu2019cyclicgen}&0.09 & 3.04 & \textcolor{blue}{\underline{35.11}} & \textcolor{blue}{\underline{0.9684}} & 32.09 & 0.9490\\
    CyclicGen\underbar{ }large\cite{liu2019cyclicgen}&- & 19.8 & 34.69 & 0.9658 & 31.46 & 0.9395\\
    DAIN\cite{bao2019dain}&0.13 &24.0 & 34.99 & 0.9683 &\textcolor{blue}{\underline{34.71}} &\textcolor{blue}{\underline{0.9756}}\\
    BMBC (Ours)&0.77 & 11.0 & \textcolor{red}{\bf 35.15} & \textcolor{red}{\bf 0.9689} & \textcolor{red}{\bf 35.01} & \textcolor{red}{\bf 0.9764}\\[-0.1em]
    \bottomrule\\[-2em]
    \end{tabular}
    }
    \label{table:Evaluation on test set}
\end{table*}

\begin{figure*}[t]
    \setlength{\belowcaptionskip}{-10pt}
	\centering
	
	\subfloat {\includegraphics[width=1.67cm]{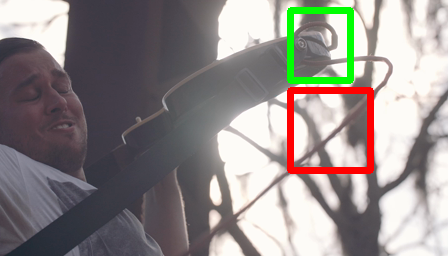}}\!
	\subfloat {\includegraphics[width=1.67cm]{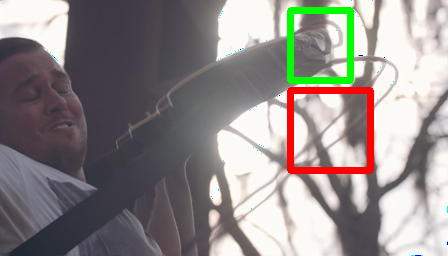}}\!
	\subfloat {\includegraphics[width=1.67cm]{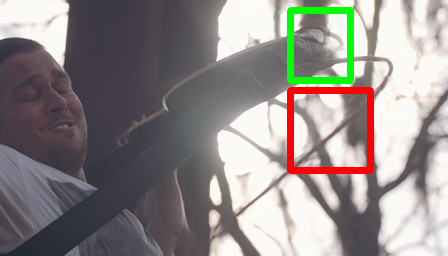}}\!
	\subfloat {\includegraphics[width=1.67cm]{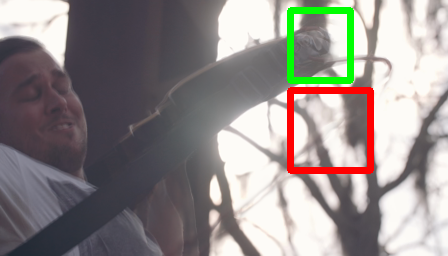}}\!
	\subfloat {\includegraphics[width=1.67cm]{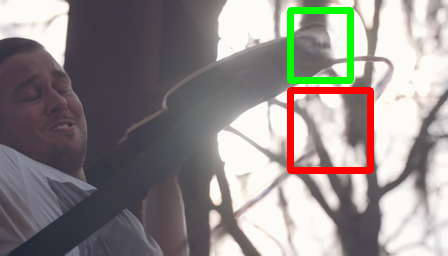}}\! %MEMC
	\subfloat {\includegraphics[width=1.67cm]{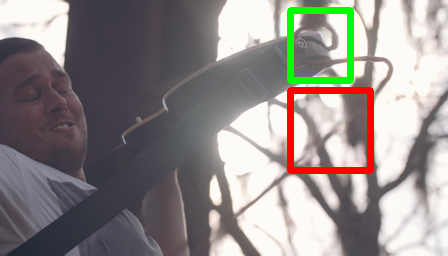}}\!
	\subfloat {\includegraphics[width=1.67cm]{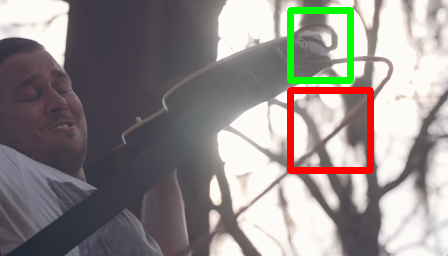}}\\[-1.1em]

	\subfloat {\includegraphics[width=1.67cm]{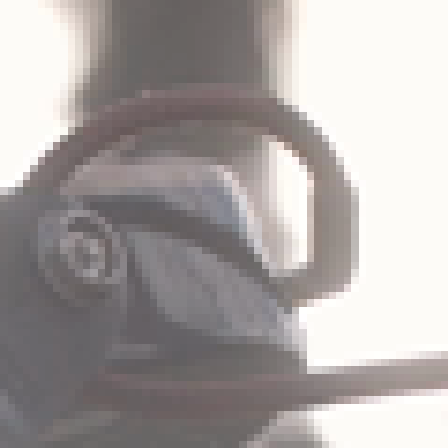}}\!
	\subfloat {\includegraphics[width=1.67cm]{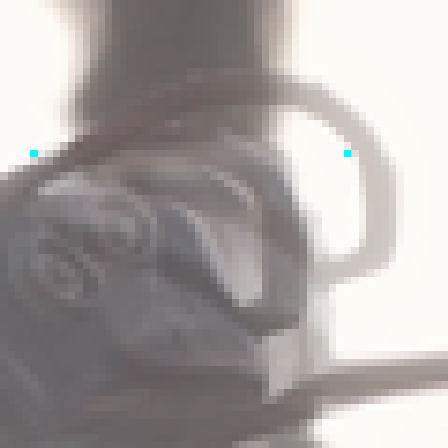}}\!
	\subfloat {\includegraphics[width=1.67cm]{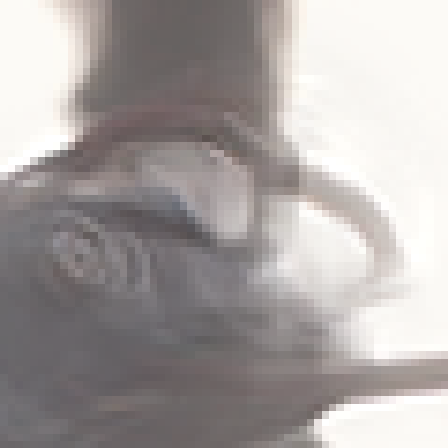}}\!
	\subfloat {\includegraphics[width=1.67cm]{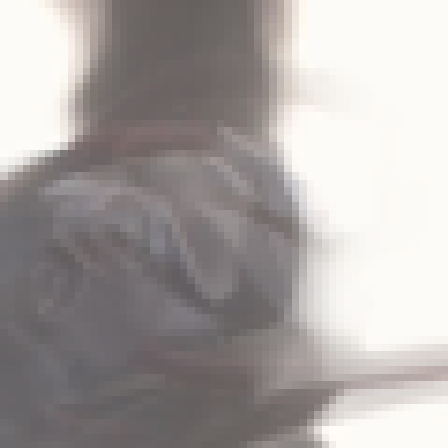}}\!
	\subfloat {\includegraphics[width=1.67cm]{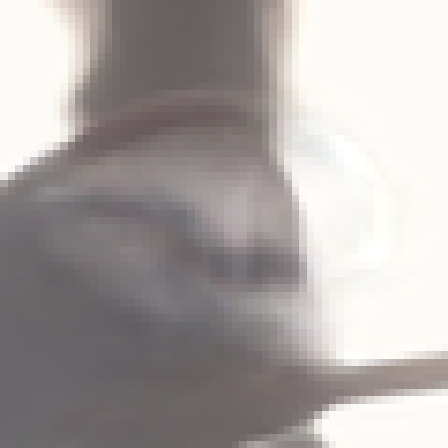}}\! %MEMC
	\subfloat {\includegraphics[width=1.67cm]{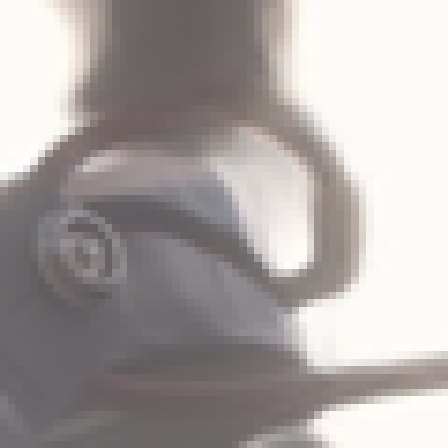}}\!
	\subfloat {\includegraphics[width=1.67cm]{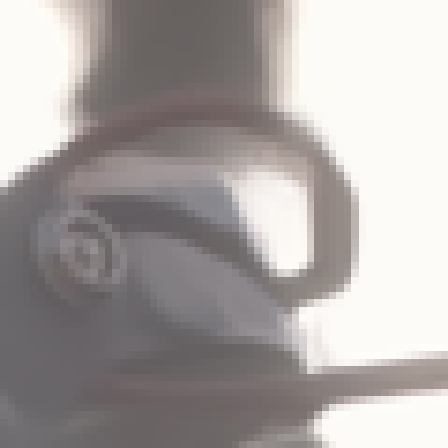}}\\[-1.1em]

    \setcounter{subfigure}{0}
	\subfloat [] {\includegraphics[width=1.67cm]{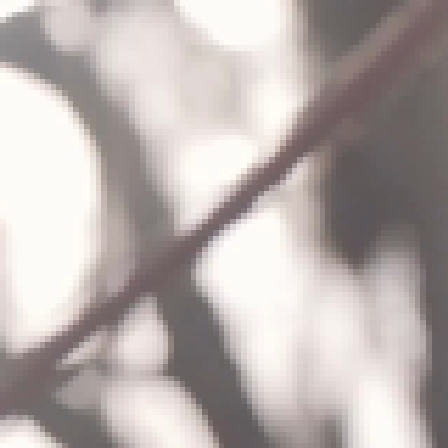}}\!
	\subfloat [] {\includegraphics[width=1.67cm]{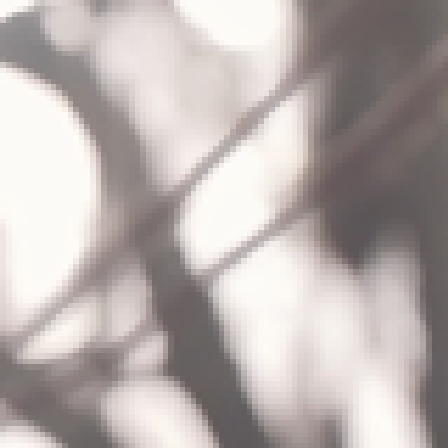}}\!
	\subfloat [] {\includegraphics[width=1.67cm]{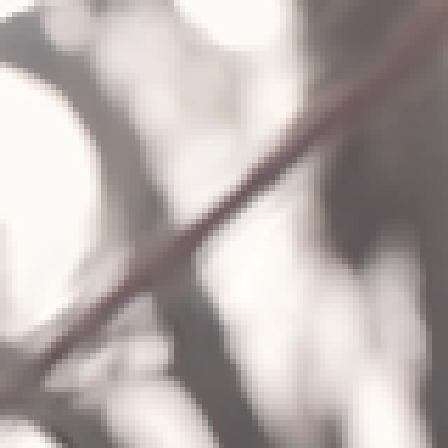}}\!
	\subfloat [] {\includegraphics[width=1.67cm]{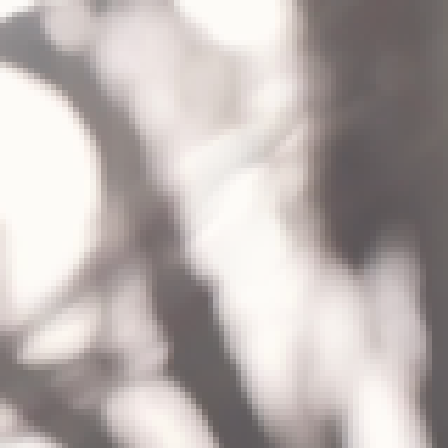}}\!
	\subfloat [] {\includegraphics[width=1.67cm]{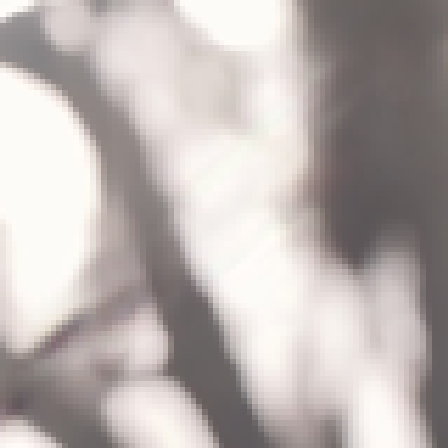}}\! %MEMC
	\subfloat [] {\includegraphics[width=1.67cm]{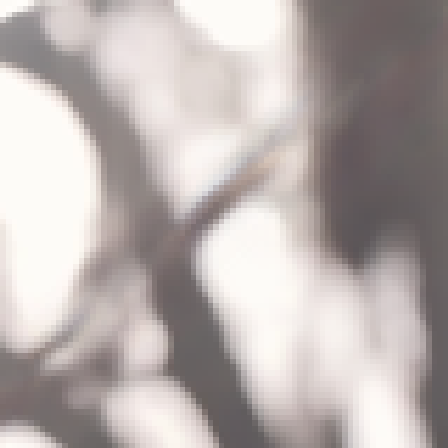}}\!
	\subfloat [] {\includegraphics[width=1.67cm]{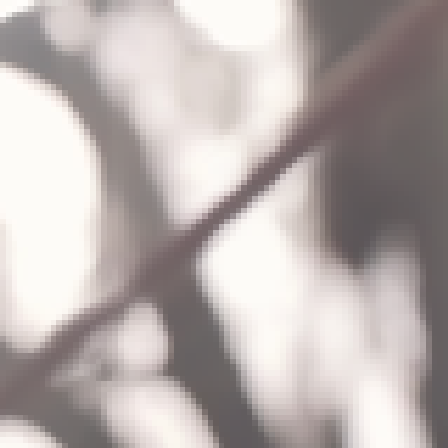}}\\[-0.8em]
    \caption
	{
        Visual comparison on the Viemo90K test set. (a) Ground-truth, (b) ToFlow~\cite{xue2019toflow}, (c) SepConv-$L_f$~\cite{niklaus2017sepconv}, (d) CyclicGen~\cite{liu2019cyclicgen}, (e) MEMC-Net*~\cite{bao2018memc}, (f) DAIN~\cite{bao2019dain}, and (g) BMBC (Ours).
	}
	\label{fig:Eval on Vimeo90K}
\end{figure*}

In Table~\ref{table:Evaluation on test set}, we provide quantitative comparisons on the UCF101~\cite{soomro2012ucf} and Vimeo90K~\cite{xue2019toflow} datasets. We compute the average PSNR and SSIM scores. The proposed algorithm outperforms the conventional algorithms by significant margins. Especially, the proposed algorithm provides 0.3 dB higher PSNR than DAIN~\cite{bao2019dain} on Vimeo90K. Fig.~\ref{fig:Eval on Vimeo90K} compares interpolation results qualitatively. Because the cable moves rapidly and the background branches make the motion estimation difficult, all the conventional algorithms fail to reconstruct the cable properly. In contrast, the proposed algorithm faithfully interpolates the intermediate frame, providing fine details.

\begin{table*}[t]
    \caption
    {
        Quantitative comparisons on the Adobe240-fps dataset for $\times2$, $\times4$, and $\times8$ frame interpolation.
    }
    \centering
    {\scriptsize
    \begin{tabular}{L{2.5cm}C{1cm}C{1cm}C{1cm}C{1cm}C{1cm}C{1cm}}
    \toprule
    \multirow{2}[2]{*}{} & \multicolumn{2}{c}{$\times2$} & \multicolumn{2}{c}{$\times4$} & \multicolumn{2}{c}{$\times8$}\\[-0.1em]
    \cmidrule(lr){2-3} \cmidrule(lr){4-5} \cmidrule(lr){6-7}
    & PSNR & SSIM & PSNR & SSIM & PSNR & SSIM\\[-0.1em]
    \midrule

    ToFlow\cite{xue2019toflow}&28.51&0.8731&29.20&0.8807&28.93&0.8812\\
    SepConv-$L_f$\cite{niklaus2017sepconv} &29.14&0.8784&29.75&0.8907&30.07&0.8956\\
    SepConv-$L_1$\cite{niklaus2017sepconv} &29.31&0.8815&\textcolor{blue}{\underline{29.91}}&\textcolor{blue}{\underline{0.8935}}&30.23&\textcolor{blue}{\underline{0.8985}}\\
    CyclicGen\cite{liu2019cyclicgen} &\textcolor{blue}{\underline{29.39}}&0.8787&29.72&0.8889&30.18& 0.8972\\
    CyclicGen\underbar{ }large\cite{liu2019cyclicgen}&28.90&0.8682&29.70&0.8866&\textcolor{blue}{\underline{30.24}}&0.8955 \\
    DAIN\cite{bao2019dain} &29.35&\textcolor{blue}{\underline{0.8820}}&29.73&0.8925&30.03&0.8983\\
    BMBC (Ours) &\textcolor{red}{\bf 29.49}&\textcolor{red}{\bf 0.8832}&\textcolor{red}{\bf 30.18}&\textcolor{red}{\bf 0.8964}&\textcolor{red}{\bf 30.60}&\textcolor{red}{\bf 0.9029}\\[-0.1em]
    \bottomrule\\[-2em]
    \end{tabular}
    }
    \label{table:Multi frame Evaluation}
\end{table*}

\begin{figure*}[t]
    \setlength{\belowcaptionskip}{-10pt}
	\centering
	
	\subfloat {\includegraphics[width=1.67cm]{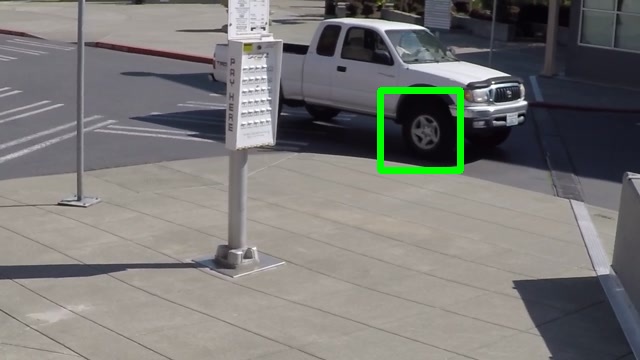}}\!
	\subfloat {\includegraphics[width=1.67cm]{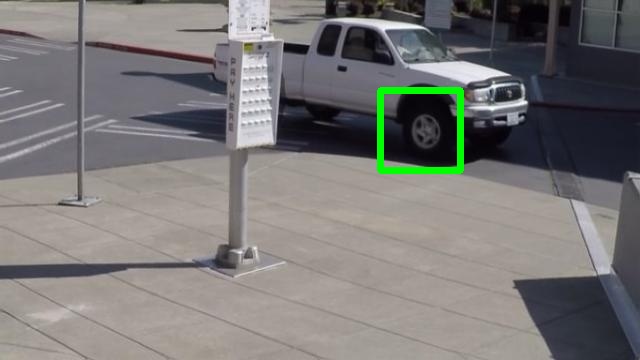}}\!
	\subfloat {\includegraphics[width=1.67cm]{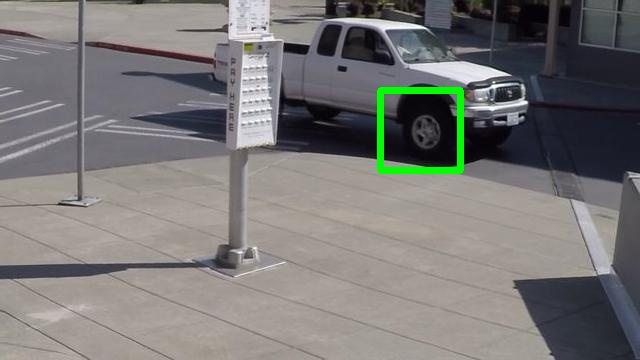}}\!
	\subfloat {\includegraphics[width=1.67cm]{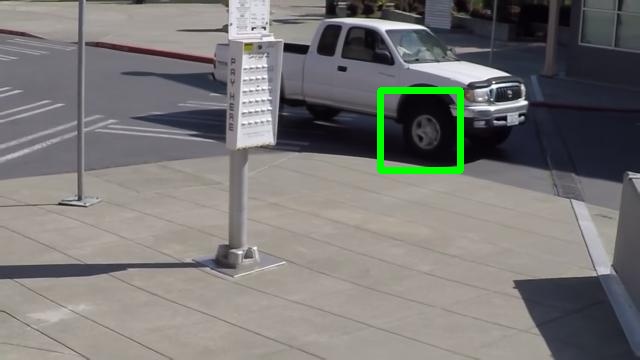}}\! 	
    \subfloat {\includegraphics[width=1.67cm]{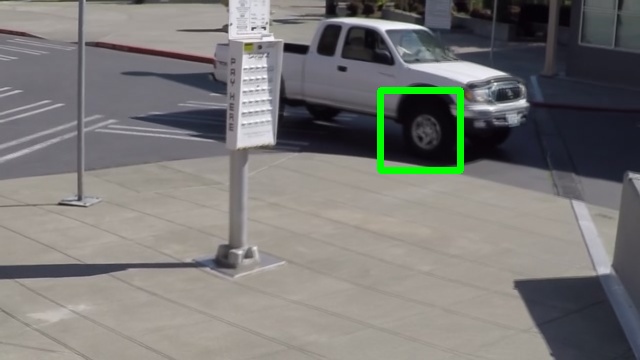}}\!
	\subfloat {\includegraphics[width=1.67cm]{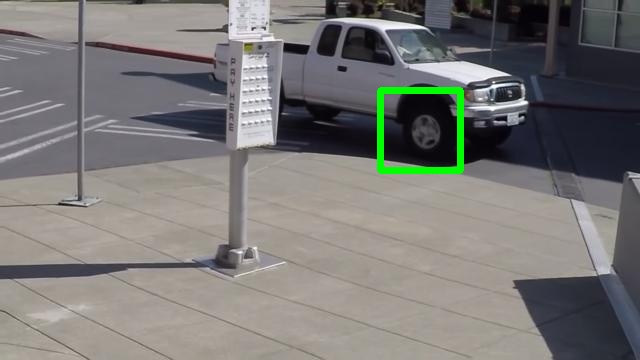}}\!
	\subfloat {\includegraphics[width=1.67cm]{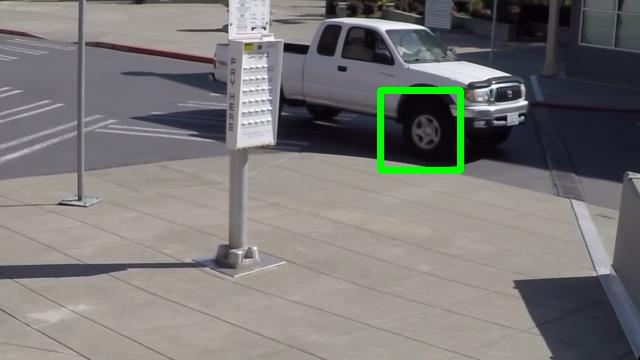}}\\[-1em]
	
    \setcounter{subfigure}{0}
	\subfloat [] {\includegraphics[width=1.67cm]{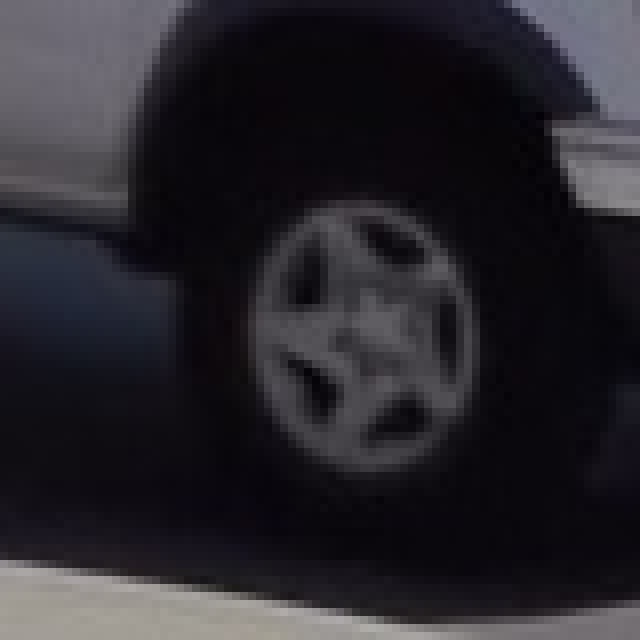}}\!
	\subfloat [] {\includegraphics[width=1.67cm]{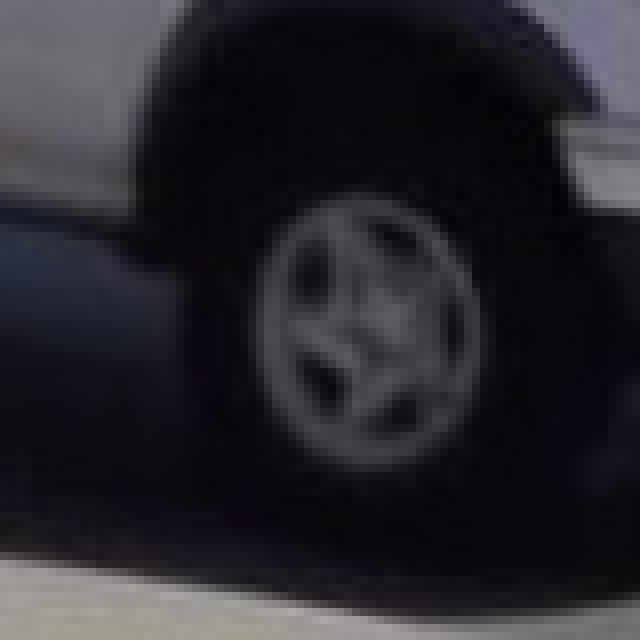}}\!
	\subfloat [] {\includegraphics[width=1.67cm]{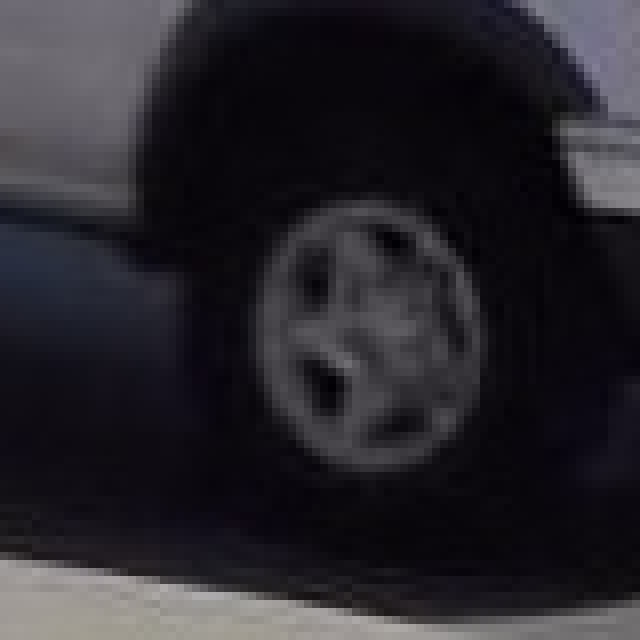}}\!
	\subfloat [] {\includegraphics[width=1.67cm]{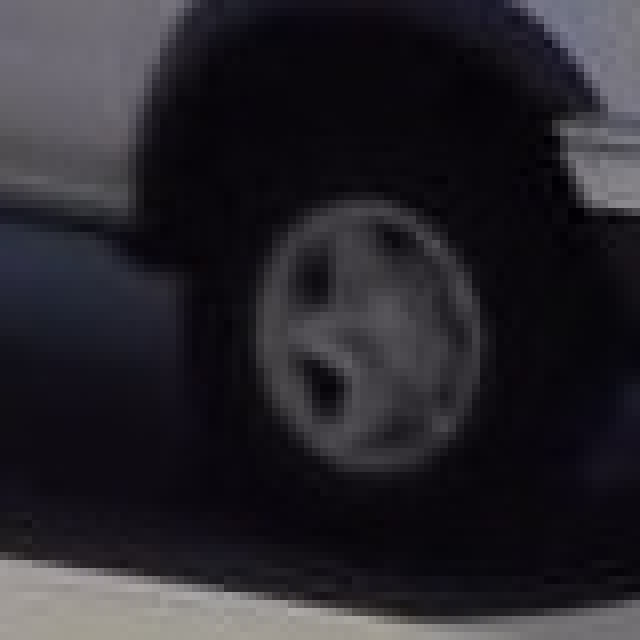}}\!
	\subfloat [] {\includegraphics[width=1.67cm]{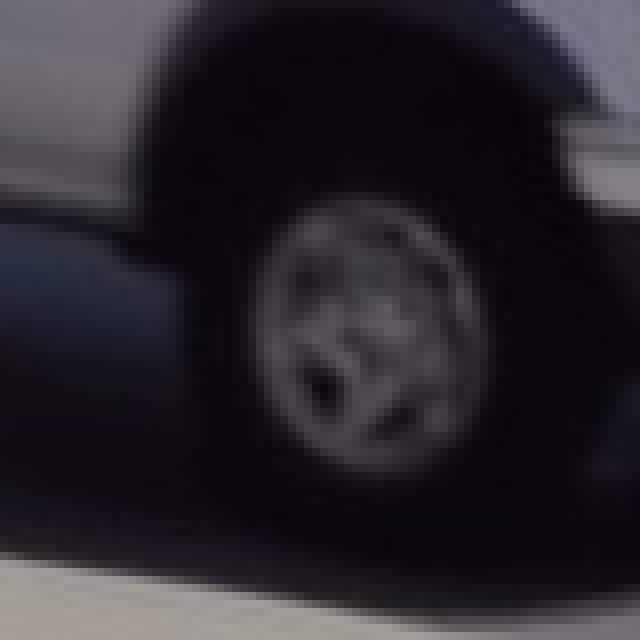}}\!
	\subfloat [] {\includegraphics[width=1.67cm]{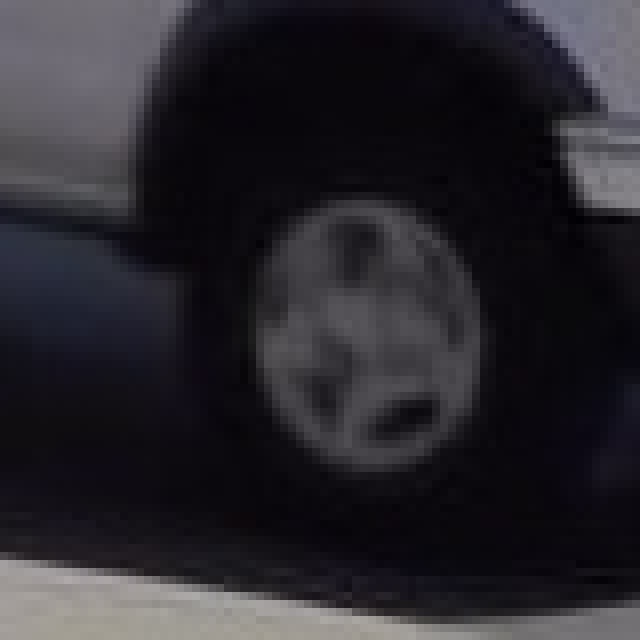}}\!
	\subfloat [] {\includegraphics[width=1.67cm]{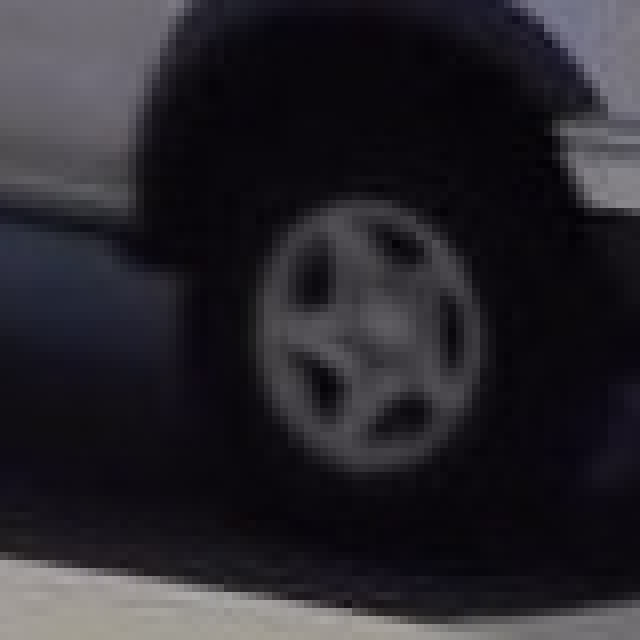}}\\[-0.8em]
    \caption
	{
        Visual comparison on the Adobe240-fps dataset. (a) Ground-truth, (b) ToFlow~\cite{xue2019toflow}, (c) SepConv-$L_f$~\cite{niklaus2017sepconv}, (d) SepConv-$L_1$\cite{niklaus2017sepconv}, (e) CyclicGen~\cite{liu2019cyclicgen}, (f) DAIN~\cite{bao2019dain}, and (g) BMBC (Ours).
	}
	\label{fig:Eval on Adobe}
\end{figure*}

The proposed algorithm can interpolate an intermediate frame at any time instance $t \in (0,1)$. To demonstrate this capability, we assess the $\times2$, $\times4$, and $\times 8$ frame interpolation performance on the Adobe240-fps dataset~\cite{Su2017adobe}. Because the conventional algorithms in Table~\ref{table:Multi frame Evaluation}, except for DAIN~\cite{bao2019dain}, can generate only intermediate frames at $t = 0.5$, we recursively apply those algorithms to interpolate intermediate frames at other $t$'s. Table~\ref{table:Multi frame Evaluation} shows that the proposed algorithm outperforms all the state-of-the-art algorithms. As the frame rate increases, the performance gain of the proposed algorithm against conventional algorithms gets larger. In Fig.~\ref{fig:Eval on Adobe}, the proposed algorithm reconstructs the details in the rotating wheel more faithfully than the conventional algorithms.

\subsection{Model analysis}

We conduct ablation studies to analyze the contributions of the three key components in the proposed algorithm: bilateral cost volume, intermediate motion approximation, and dynamic filter generation network. By comparing various combinations of intermediate candidates, we analyze the efficacy of the bilateral cost volume and the intermediate motion approximation jointly.

\begin{table*}[t]
    \caption
    {
        PSNR comparison of combination of the intermediate candidates.
    }
    \centering
    \scriptsize
    \begin{tabular}{L{3.0cm}C{2.1cm}C{2.1cm}}
    \toprule
    \multirow{2}[1]{*}{Intermediate candidates}& UCF101~\cite{soomro2012ucf} & Vimeo90K~\cite{xue2019toflow}\\[-0.1em]
    \cmidrule(lr){2-2} \cmidrule(lr){3-3}
    & PSNR & PSNR\\[-0.1em]
    \midrule
    Appx4&  34.99&34.72\\
    BM&  35.12&34.93\\
    BM+Appx2&  35.14&34.95\\
    BM+Appx4& 35.15 &35.01\\[-0.1em]
    \bottomrule
    \end{tabular}
    \label{table:Intermeidate candidates}
\end{table*}

\subsubsection{Intermediate candidates:}
To analyze the effectiveness of the bilateral motion estimation and the intermediate motion approximation, we train the proposed networks to synthesize intermediate frames using the following combinations:
\begin{itemize}
    \setlength\itemsep{0.5em}
    \resitem{Appx4: Four intermediate candidates, obtained using approximated bilateral motions in \eqref{eq:app_forward_t1}$\sim$\eqref{eq:app_backward_t1}, are combined.}
    \resitem{BM: Two intermediate candidates, obtained using bilateral motions, are combined.}
    \resitem{BM+Appx2: In addition to BM, two more candidates  obtained using approximated bilateral motions  $V^{fw}_{t \rightarrow 0}$ in \eqref{eq:app_forward_t0} and $V^{bw}_{t \rightarrow 1}$ in \eqref{eq:app_backward_t1} are used.}
    \resitem{BM+Appx4: Six intermediate candidates are used as well (proposed model).}
\end{itemize}

Table~\ref{table:Intermeidate candidates} compares these models quantitatively.
First, Appx4 shows the worst performance, while it is still comparable to the state-of-the-art algorithms. Second, BM provides better performance than Appx4 as well as the state-of-the-art algorithms, which confirms the superiority of the proposed BM to the approximation. Third, we can achieve even higher interpolation performance with more intermediate candidates obtained through the motion approximation.

\subsubsection{Dynamic blending filters}

\begin{table*}[t]
    \caption
    {
        Analysis of the dynamic filter generation network. In all settings, the six warped frames are input to the network.
    }
    \centering
    \scriptsize
    \begin{tabular}{C{1.8cm}C{2.1cm}C{2.1cm}C{2.1cm}C{2.1cm}}
    \toprule
    \multirow{2}[1]{*}{Kernel size}&\multicolumn{2}{c}{Input to filter generation network}& UCF101~\cite{soomro2012ucf} & Vimeo90K~\cite{xue2019toflow}\\[-0.1em]
    \cmidrule(lr){2-3} \cmidrule(lr){4-4} \cmidrule(lr){5-5}
     & Input frames & Context maps & PSNR & PSNR\\[-0.1em]
    \midrule
    $5\times5$  &  &  & 34.98 & 34.81\\
    $3\times3$  & \checkmark &  & 35.09 & 34.90\\
    $5\times5$  & \checkmark &  & 35.08 & 34.96\\
    $7\times7$  & \checkmark &  & 35.02 & 34.98\\
    $5\times5$  & \checkmark & \checkmark & 35.15 & 35.01\\[-0.1em]
    \bottomrule\\[-2em]
    \end{tabular}
    \label{table:ablation dynamic filter}
\end{table*}

We analyze the optimal kernel size and the input to the dynamic filter generation network. Table~\ref{table:ablation dynamic filter} compares the PSNR performances of different settings. First, the kernel size has insignificant impacts, although the computational complexity is proportional to the kernel size. Next, when additional information (input frames and context maps) is fed into the dynamic filter generation network, the interpolation performance is improved. More specifically, using input frames improves PSNRs by 0.10 and  0.15 dB on the UCF101 and Vimeo90K datasets. Also, using context maps further improves the performances by 0.07 and 0.05dB. This is because the input frames and context maps help restore geometric structure and exploit rich contextual information.

\section{Conclusions}

We developed a deep-learning-based video interpolation algorithm based on the bilateral motion estimation, which consists of the bilateral motion network and the dynamic filter generation network. In the bilateral motion network, we developed the bilateral cost volume to estimate accurate bilateral motions. In the dynamic filter generation network, we warped the two input frames using the estimated bilateral motions and fed them to learn filter coefficients. Finally, we synthesized the intermediate frame by superposing the warped frames with the generated blending filters. Experimental results showed that the proposed algorithm outperforms the state-of-the-art video interpolation algorithms on four benchmark datasets.

\section*{Acknowledgements}
This work was supported in part by the Agency for Defense Development (ADD) and Defense Acquisition Program Administration (DAPA) of Korea under grant UC160016FD and in part by the National Research Foundation of Korea (NRF) grant funded by the Korea Government (MSIP) (No. NRF-2018R1A2B3003896 and No. NRF-2019R1A2C4069806).

\clearpage
\bibliographystyle{splncs04}
\bibliography{2095}
\end{document}